\documentclass{article}

\usepackage{microtype}
\usepackage{graphicx}
\usepackage{subfigure}
\usepackage{booktabs} 

\usepackage{hyperref}
\usepackage{xspace}

\newcommand{\bandwidth}{\varphi}
\newcommand{\edgeset}{\mathcal{E}}
\newcommand{\cmbandwidth}{\hat \varphi}
\newcommand{\bigO}{\mathcal{O}}

\newcommand{\shortmethodname}{\mbox{BwR}\@\xspace}
\newcommand{\ginestack}{\texttt{GINEStack}\@\xspace}

\usepackage[accepted]{icml2023}

\usepackage{amsmath}
\usepackage{amssymb}
\usepackage{mathtools}
\usepackage{amsthm}

\usepackage{colortbl}
\usepackage{multirow}
\definecolor{gg}{gray}{0.92}
\newcolumntype{a}{>{\columncolor{gg}}c}

\usepackage[capitalize,noabbrev]{cleveref}

\usepackage{placeins}

\theoremstyle{plain}

\theoremstyle{definition}

\theoremstyle{remark}

\usepackage[textsize=tiny]{todonotes}

\icmltitlerunning{Bandwidth-Restricted Graph Generation}

\begin{document}

\twocolumn[
\icmltitle{Improving Graph Generation by Restricting Graph Bandwidth}



\icmlsetsymbol{equal}{*}

\begin{icmlauthorlist}
\icmlauthor{Nathaniel Diamant}{genentech}
\icmlauthor{Alex M. Tseng}{genentech}
\icmlauthor{Kangway V. Chuang}{genentech}
\icmlauthor{Tommaso Biancalani}{genentech}
\icmlauthor{Gabriele Scalia}{genentech}
\end{icmlauthorlist}

\icmlaffiliation{genentech}{Department of Artificial Intelligence and Machine Learning, Research and Early Development, Genentech, USA}

\icmlcorrespondingauthor{Nathaniel Diamant}{diamant.nathaniel@gene.com}
\icmlcorrespondingauthor{Gabriele Scalia}{scalia.gabriele@gene.com}

\icmlkeywords{graph generation, graph theory, deep generative modeling, graph neural network, graph bandwidth}

\vskip 0.3in
]



\printAffiliationsAndNotice{}  

\begin{abstract}

Deep graph generative modeling has proven capable of learning the distribution of complex, multi-scale structures characterizing real-world graphs. However, one of the main limitations of existing methods is their large output space, which limits generation scalability and hinders accurate modeling of the underlying distribution. To overcome these limitations, we propose a novel approach that significantly reduces the output space of existing graph generative models. Specifically, starting from the observation that many real-world graphs have low graph bandwidth, we restrict graph bandwidth during training and generation. Our strategy improves both generation scalability and quality without increasing architectural complexity or reducing expressiveness. Our approach is compatible with existing graph generative methods, and we describe its application to both autoregressive and one-shot models. We extensively validate our strategy on synthetic and real datasets, including molecular graphs. Our experiments show that, in addition to improving generation efficiency, our approach consistently improves generation quality and reconstruction accuracy.
The implementation is made available\footnote{\url{https://github.com/Genentech/bandwidth-graph-generation}}.

\end{abstract}

\section{Introduction} \label{sec:introduction}

Learning the underlying distribution of graphs for generative purposes finds important applications in diverse fields, where objects can be naturally described through their structures~\cite{DBLP:journals/debu/HamiltonYL17}. Computational approaches to capture graph statistical properties have long been established~\cite{albert2002statistical}, whereas deep generative modeling has recently been proven capable of learning both global and fine-grained structural properties, along with their complex interdependencies \cite{guo2022systematic}. These results suggest many promising applications for deep graph generative modeling, which include the biomedical and pharmaceutical domains, where essential objects such as molecules, gene networks, and cell-level tissue organization can be represented as graphs~\cite{li2022graph}. Despite these promises, several open challenges remain. 

Applications in these domains require modeling graphs with a high number of nodes $N$ that leads to a large output space, where the number of possible edges is in $\bigO(N^2)$.
At the same time, many real-world graphs are sparse, and they are characterized by a small number of semantically rich connections which need to be accurately modeled (e.g., a small subset of chemical bonds can confer radically different properties in a molecular graph).

Recent research has focused on accurately learning complex dependencies leveraging generative models such as variational autoencoders (VAEs) \cite{pmlr-v97-grover19a}, recurrent neural networks (RNNs) \cite{pmlr-v80-you18a}, normalizing-flow models \cite{Shi*2020GraphAF:} and score-based models \cite{pmlr-v108-niu20a}. A general limitation of these approaches is their high time complexity and output space $\mathcal{O}(N^2)$.
This limits both their scalability and makes accurate prediction of sparse connections challenging, as the ratio of observed to possible edges can be extremely small.

For this reason, more tractable methods have been proposed, which leverage different architectures~\cite{li2018learning,liu2018constrained,dai2020scalable}, generate coarse-grained motifs \cite{jin2018junction,liao2019efficient}, or change the output representation, such as transforming graphs to sequences \cite{goyal2020graphgen} or using domain-specific encodings such as molecular SMILES~\cite{gomez2018automatic}. Although these approaches are more scalable, they trade-off efficiency with model complexity, expressiveness, or have limited applicability because of domain-specific choices. 

To overcome the limitations of existing approaches, we propose a novel strategy: \shortmethodname (\emph{Bandwidth-Restricted}) graph generation. \shortmethodname leverages a permutation of the adjacency matrix to restrict the \emph{graph bandwidth}, reducing both the time complexity and the output space from $\mathcal{O}(N^2)$ to $\mathcal{O}(N\cdot \cmbandwidth)$, where $\cmbandwidth$ is the estimated bandwidth. As will be shown, $\cmbandwidth$ is low for many classes of real-world graphs, such as those characterizing the biomedical domain. During training, \shortmethodname leverages bandwidth-restricted adjacency matrices; during sampling, it constrains the generation within a bandwidth-restricted space, which reduces both time complexity and output space without losing expressiveness.

This strategy brings two key advantages. First, reducing  time complexity improves \emph{generation scalability} (i.e., time and memory requirements). Second, reducing the output space simplifies learning the underlying data distribution, while also making the ratio of observed to possible edges less imbalanced, with a  positive impact on \emph{generation quality}.
Importantly, \shortmethodname can be easily integrated into virtually all existing  graph generative methods, as it is orthogonal to the generative model architecture. Therefore, it does not increase model complexity nor add domain-specific constraints.
We describe our strategy in the context of three recent models: an autoregressive model based on GraphRNN \cite{pmlr-v80-you18a}, a one-shot VAE-based model based on Graphite  \cite{pmlr-v97-grover19a}, and a one-shot score-based model based on EDP-GNN \cite{pmlr-v108-niu20a}.

We experimentally validate \shortmethodname by evaluating the reconstruction accuracy and generation quality on both synthetic and real datasets. Additionally, we analyze memory and time improvements. We include molecular datasets---spanning both small molecules and larger peptides---to evaluate the advantages of \shortmethodname for \textit{de novo} molecular generation. Our results show that \shortmethodname consistently achieves superior or competitive generative performance to the standard baselines, at a fraction of the time/space complexity.

\textbf{Contributions.} We summarize our main contributions as follows:

\begin{itemize}
    \item We show that many real-world classes of graphs, such as molecules, have low graph bandwidth.
    \item Building on this property, we propose  \emph{\shortmethodname} (Bandwidth-Restricted) graph generation, a novel strategy for graph generation that constrains the bandwidth, drastically reducing time complexity and output space. 
    \item \shortmethodname can be applied to virtually all existing graph generation methods. We describe its application to an autoregressive method, GraphRNN \cite{pmlr-v80-you18a}; a one-shot VAE-based method, Graphite \cite{pmlr-v97-grover19a}; and a one-shot score-based method, EDP-GNN \cite{pmlr-v108-niu20a}.
    \item We validate \shortmethodname on both synthetic and real-world datasets, with a focus on real-world molecular datasets. Our results show that, in addition to being more efficient in terms of time and memory used, \shortmethodname consistently improves reconstruction accuracy and generative quality across datasets and methods. 
\end{itemize}

\begin{figure*}[ht]
\vskip 0.2in
\begin{center}
\centerline{
\includegraphics[width=2\columnwidth]{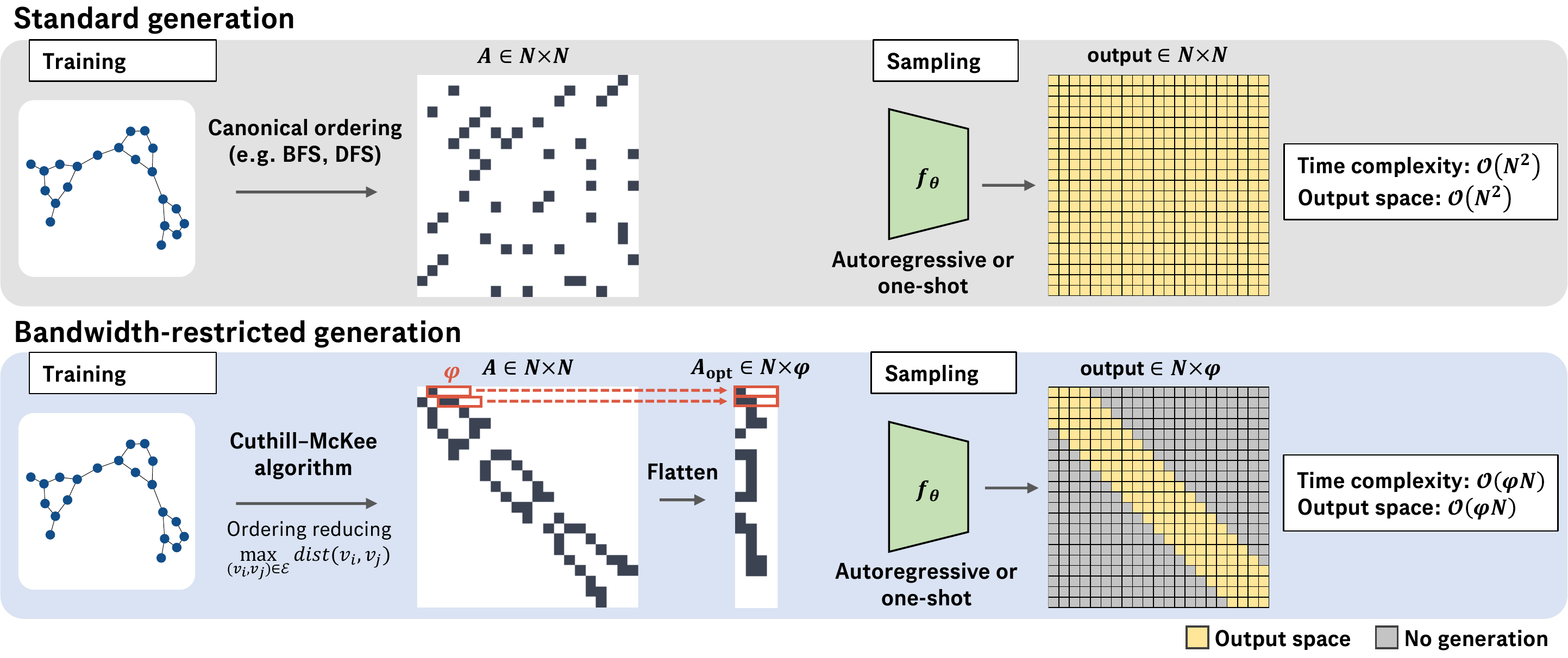}
}
\caption{Overview of our strategy and comparison with standard generation methods.  \textbf{(Top)} In a standard graph generative method, the model is trained on adjacency matrices $A$ derived through a specific canonical ordering on the graph (e.g., BFS or DFS). During sampling, the model needs to predict edges from a space in $\bigO\left(N^2\right)$. \textbf{(Bottom)} Our bandwidth-restricted graph generation leverages the Cuthill-Mckee (C-M) ordering \citep{cuthill1969reducing} to reduce the bandwidth $\bandwidth(A)$ of each adjacency matrix.
The C-M order results in an adjacency matrix that is a \emph{band matrix}, with all-zero entries outside a $\bandwidth(A)$-sized band. $A$ is re-parameterized as $A^{\text{opt}} \in N \times \bandwidth(A)$, which is used for training. During sampling, only edges in an $N \times \bandwidth(A)$ space (yellow) are considered as candidates, thus drastically reducing the output space to $\bigO\left(N \times \bandwidth(A)\right)$.}
\label{fig:overview}
\end{center}
\vskip -0.2in
\end{figure*}

\section{Related Work} \label{sec:related}

\subsection{Graph Generative Models}
Graph generative models seek to learn the underlying distribution of graph datasets. A model is trained on a set of observed graphs $\mathcal{G} = \left\{ G_1, \ldots, G_S \right\} \sim p\left(G\right)$, where each graph $G_i = \left(\mathcal{V}_i, \mathcal{E}_i\right)$ is defined by its set of nodes $\mathcal{V}_i = \left(v_1, \ldots, v_N \right)$ and edges $\mathcal{E}_i \subseteq \mathcal{V}_i \times \mathcal{V}_i$. The model learns the distribution $p_{\text{model}}\left(G\right) \approx p\left(G\right)$ that allows sampling new graphs.
Broadly, graph generative models can be categorized as \emph{autoregressive} \cite{pmlr-v80-you18a,Shi*2020GraphAF:,goyal2020graphgen,li2018learning,liu2018constrained} or \emph{one-shot} \cite{https://doi.org/10.48550/arxiv.1611.07308, ma2018constrained, pmlr-v97-grover19a, pmlr-v108-niu20a} models. A common way to represent the graph topology is through its adjacency matrix $A^\pi \in N \times N$. The adjacency matrix depends on a specific node ordering $\pi$, defined as a bijective function~\mbox{$\pi : \mathcal{V} \rightarrow  \left[ 1, N \right]$}. 

Autoregressive models treat graph generation as a sequential decision process, factorizing $p_{\text{model}}\left(G\right)$ into the joint probability of its components (e.g., nodes or motifs). Node-based autoregressive methods \cite{pmlr-v80-you18a,Shi*2020GraphAF:} generate each node and its edges in a predefined order, conditioning each new node on the already-generated graph, with time complexity and output space in $\mathcal{O}(N^2)$.  Solutions have been proposed to separately output actions corresponding to the number of new edges and their identities for each new node, reducing time complexity 
\cite{li2018learning,liu2018constrained}. However, these methods increase model complexity without reducing the output space. In contrast, \shortmethodname reduces \emph{both} the time complexity and output space of existing node-based autoregressive models to be in $\mathcal{O}(N\cdot \cmbandwidth)$. This reduction is achieved without losing expressiveness and without increasing model complexity. 

One-shot models sample the whole topology of the graph from a latent distribution.
Many one-shot models use permutation-invariant functions to output the graph topology. This class of generative models can be split into two main categories: adjacency-matrix-based models \cite{ma2018constrained, pmlr-v108-niu20a} directly output $A^\pi$, while node-embedding-based models \cite{https://doi.org/10.48550/arxiv.1611.07308,pmlr-v97-grover19a} sample node embeddings from the latent distribution and compute $A^\pi$ based on pairwise relationships between them. In both cases, these methods have time complexity and output space in $\mathcal{O}(N^2)$, as they need to consider edges between every pair of nodes. In contrast, \shortmethodname allows reducing both the time complexity and output space to be in $\mathcal{O}(N \cdot \cmbandwidth)$, with no loss of expressiveness.

Last, we notice how our approach is orthogonal and compatible with other methods proposed to increase GNN efficiency---such as graph partitioning \citep{jia2020improving}---as it is largely independent of the generative method. For the present work, we focus on node-based generation, but our approach can be extended to coarser motif-based generation  \cite{liao2019efficient}.

\subsection{Graph Ordering}

A unique challenge of graph generative models is that the set of all possible orderings leads to up to $N!$ different adjacency matrices for the same graph 
\cite{liao2019efficient}.
Given that our method imposes a specific node ordering, it is related to other works that have investigated ordering in graph generation. 

Ordering is crucial in autoregressive models. For any particular ordering $\pi$, if the index distance $\left|\pi\left(v_i\right) - \pi\left(v_j\right)\right|$  between two connected nodes $(v_i, v_j) \in \mathcal{E}$  is high, the model is required to handle long-term dependencies.  This issue can be addressed by choosing a specific canonical ordering. For example, GraphRNN \cite{pmlr-v80-you18a} is trained using random breadth-first-search (BFS) node orderings for each graph, such that new nodes can only be connected to existing nodes at the frontier of the BFS. GraphRNN and its comparison with our bandwidth-restricted version are further discussed in Section \ref{ssec:method-graphrnn}.

It has been shown that the choice of node ordering impacts graph generation for specific applications. For example, in the context of autoregressive molecular generation, BFS had clear advantages over depth-first-search (DFS), even though the latter is more often used to define a canonical order for molecular structures \cite{mercado2021exploring}.

A more fundamental issue raised by non-unique graph representations is that choosing a specific ordering $\pi$ does not rigorously correspond to maximum-likelihood estimation (MLE), thus preventing exact likelihood evaluation \cite{liao2019efficient,pmlr-v139-chen21j}. Additionally, it can make the reconstruction loss ambiguous. As discussed by \citet{liao2019efficient}, training on random orderings in a specific canonical family (e.g., BFS) optimizes a variational lower bound of the true log-likelihood tighter than any single arbitrary ordering. For autoregressive models, a tighter lower bound is derived by \citet{pmlr-v139-chen21j} by performing approximate posterior inference over the node ordering. For one-shot models, \citet{winter2021permutation} addresses reconstruction ambiguity by training a permuter model to reorder generated graphs alongside a standard encoder/decoder architecture. We notice how these works are orthogonal with respect to our contribution. Indeed, bandwidth-optimized graphs define a canonical family of node orderings, and existing methods could be used to improve likelihood estimation. This integration will be investigated in future work.

\section{Graph Bandwidth Background} \label{sec:bandwidth}

We start providing a definition of bandwidth through the graph bandwidth problem \cite{unger1998complexity}.
Intuitively, the graph bandwidth problem can be seen as placing the nodes of a graph on a line such that the ``length'' of the longest edge in the graph is minimized. The bandwidth of the graph is then simply the length of the longest edge.

Given a graph \mbox{$G = \left(\mathcal{V}, \mathcal{E}\right)$} on $N$ vertices, each ordering  \mbox{$\pi : \mathcal{V} \rightarrow  \left[ 1, N \right]$} defines a graph linearization. We define the \emph{distance} between nodes $v_i$ and $v_j$ in the ordering $\pi$ as \mbox{$\textrm{dist}_\pi\left(v_i, v_j \right) = \left|\pi\left(v_i\right) - \pi\left(v_j\right)\right|$}. The bandwidth of the ordering $\pi$ is defined as the maximum stretch of any edge on the linearization, i.e. $\bandwidth\left(\pi\right) = \max_{\left(v_i, v_j\right) \in \mathcal{E}} \textrm{dist}_\pi\left(v_i, v_j \right)$.
The bandwidth of a graph $G$ is the minimum bandwidth across all possible orderings, i.e.:
\begin{equation}
    \bandwidth\left(G\right) = \min_{\pi : \mathcal{V} \rightarrow  \left[ 1, N \right]} \bandwidth\left(\pi\right).
\label{eq:bandwidth}
\end{equation}

\begin{figure}[t]
\vskip 0.2in
\begin{center}
\centerline{\includegraphics[width=0.8\columnwidth]{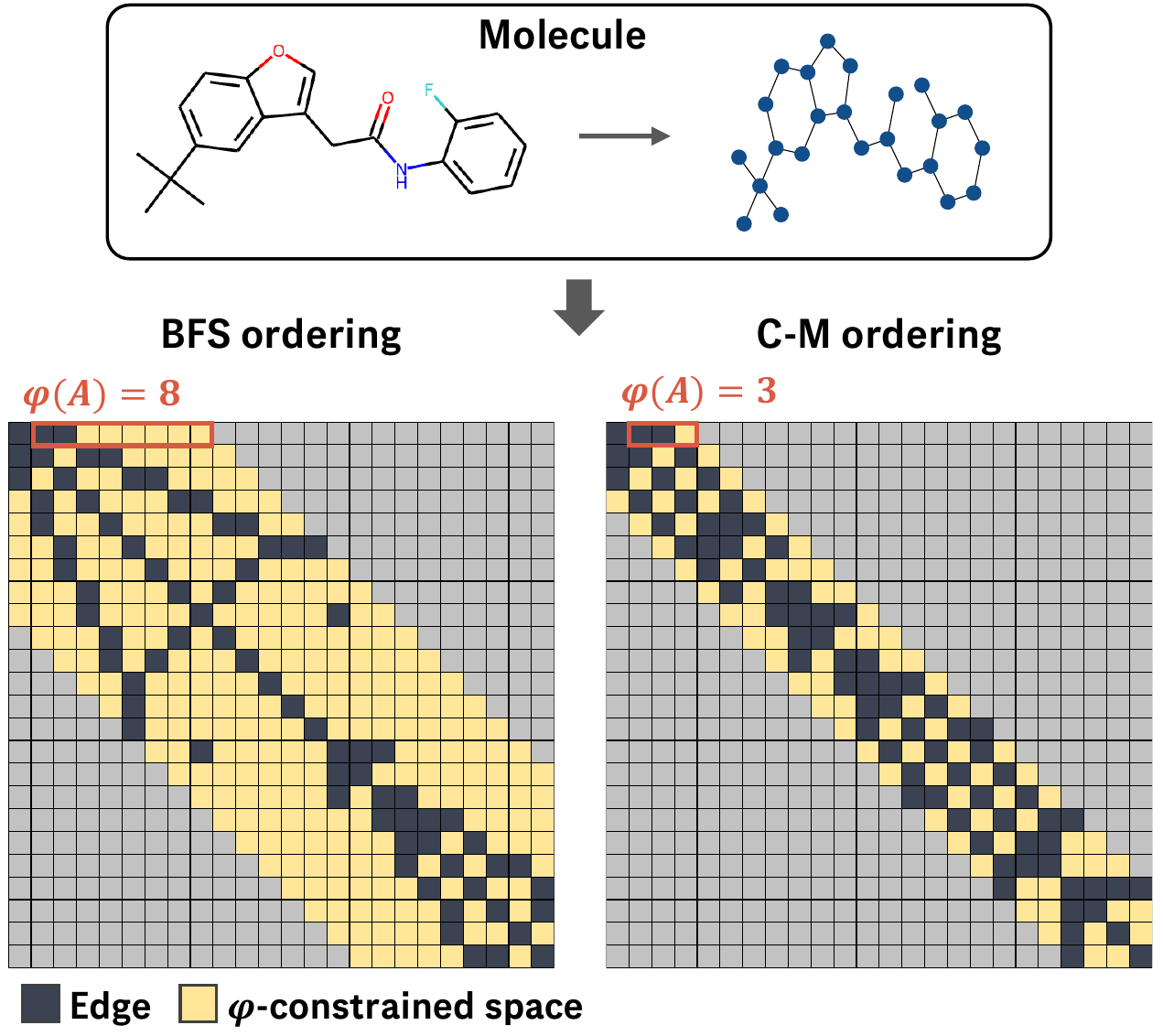}}
\caption{Bandwidth of two adjacency matrices of the same molecular graph. The left adjacency matrix is given by a BFS ordering, the right adjacency matrix is given by a Cuthill-McKee ordering.}
\label{fig:bandwidth-example}
\end{center}
\vskip -0.2in
\end{figure}

Importantly, an ordering that minimizes $\bandwidth$ results in an adjacency matrix where all non-zero entries lie in a narrow band along the diagonal (hence the term ``bandwidth''). This enables a compact matrix representation of $N\times\bandwidth$ instead of $N^2$ (Figure \ref{fig:overview}) and drastically reduces the space required to represent the graph when $\bandwidth \ll N$.
The maximum size of the off-diagonal band of the adjacency matrix is known as the matrix's bandwidth, which we denote as $\bandwidth(A)$. Figure \ref{fig:bandwidth-example} shows the bandwidth of two adjacency matrices of a molecular graph.
We note that for an ordering $\pi$, $\bandwidth(\pi) = \bandwidth(A^\pi)$.

The graph bandwidth problem has been shown to be \mbox{NP-hard} for general graphs \cite{papadimitriou1976np}, and also for simpler classes of graphs such as trees and even caterpillar trees \cite{monien1986bandwidth}.
Exact polynomial solutions exist for very restricted classes of graphs, and approximate superpolynomial solutions have been proposed for general graphs \cite{feige2000coping,cygan2010exact}. However, in practice, efficient heuristic approaches work well for general graphs and are routinely leveraged in applications.

One such heuristic is the Cuthill-McKee (C-M) algorithm (\citeyear{cuthill1969reducing}), which is based on a variation of BFS search, has linear time complexity $\mathcal{O}(|\mathcal{E}|)$  \cite{chan1980linear}, and has been extensively studied from a theoretical perspective \cite{turner1986probable}. Extensions of this algorithm and other heuristics have been proposed that improve efficiency and/or theoretical guarantees, though results are dataset dependent \cite{gonzaga2018evaluation}.
For the present work, we leverage the C-M algorithm initialized to start at a pseudo peripheral node \cite{10.2307/2156090}. However, our proposed strategy is independent of the choice of the bandwidth-minimization algorithm, and other approaches---potentially even learned---will be explored in future work. 
Interestingly, as discussed in Section~\ref{sec:related},  BFS has been shown to be superior to DFS for autoregressive molecular generation. Given that C-M can be seen as a special case of BFS, using C-M allows us to retain the benefits of BFS, while also further reducing $\max_{\left(v_i, v_j\right) \in \mathcal{E}} \textrm{dist}_\pi\left(v_i, v_j \right)$.
We define the bandwidth of the adjacency matrix derived with the C-M algorithm as $\cmbandwidth$.

A key observation that motivates the present work is that many real-world graphs, such as those characterizing the biomedical domain, have low $\cmbandwidth$. Table~\ref{tab:naturalBWs} shows the empirical bandwidth computed with the C-M algorithm for a diverse set of chemical and biological datasets (see Appendix~\ref{app:datasets} for more details on the datasets). For each dataset, a \emph{savings factor} summarizes the space reduction. The savings factor is calculated as the ratio of the number of edges in the bandwidth-restricted graph to the number of edges in the complete graph (i.e. the size of a non-bandwidth-restricted adjacency matrix). As shown, a bandwidth reparameterization leads to a savings factor $> 3$ for most small-molecule datasets (e.g., $3.8 \pm 1.0$ for ZINC250k). The savings factor is even higher for datasets including larger molecules (e.g., $13.5 \pm 6.5$ for the Peptides-func dataset). Significant savings are also confirmed on non-molecular datasets, such as brain networks (KKI, OHSU). The high savings factor of molecular datasets is due to the existence of an intrinsic upper bound on the bandwidth of molecule-like graphs, which we derive and further discuss in Appendix~\ref{app:bw-molecular-theory}.

The C-M algorithm consistently reduces $\bandwidth(\pi)$ compared to the orderings routinely used in graph generation (BFS, DFS, etc.). Figure \ref{fig:bandwidth-example} compares the adjacency matrices of a molecular graph given by BFS and C-M, and their respective bandwidths. The bandwidth decreases from 8 to 3, which translates into a two-fold reduction of the output space. Additional examples of adjacency matrices for molecular graphs given by BFS, DFS, RDKit \citep{greg-landrum-rdkit}, and \mbox{C-M} orderings are presented in Figure \ref{fig:mol-orderings} (Appendix). As shown, the C-M ordering consistently leads to the lowest $\bandwidth$. 
Overall, the C-M algorithm allows reducing the bandwidths of all the considered datasets.
For example, 95\% of the molecules in ZINC250k have \mbox{$\cmbandwidth \le 4$} (Figure \ref{fig:zincBW}, Appendix).

\begin{table}[t]
\caption{Number of nodes, C-M bandwidth, and savings factor of a set of chemical and biological datasets. The first section of the table consists of small molecules, the second section consists of large molecules, and the third section consists of brain networks. See Appendix~\ref{app:datasets} for more details on the datasets. 
Average and standard deviation calculated across graphs for each dataset.
}
\label{tab:naturalBWs}
\vskip 0.15in
\begin{center}
\begin{small}

\scalebox{0.9}{
\begin{tabular}{cccc}
\toprule
Dataset              & $N$                 & $\cmbandwidth$         & \begin{tabular}[c]{@{}c@{}}Savings\\ Factor\end{tabular}         \\ \midrule
ZINC250k & $23.2 \pm 4.5$ & $3.3 \pm 0.8$ & $3.9 \pm 1.0$ \\
AIDS & $14.0 \pm 10.4$ & $3.1 \pm 1.1$ & $2.4 \pm 1.0$ \\
Alchemy & $10.1 \pm 0.7$ & $2.8 \pm 0.7$ & $2.1 \pm 0.4$ \\
MCF-7 & $26.1 \pm 10.7$ & $4.1 \pm 1.3$ & $3.5 \pm 1.2$ \\
MOLT-4 & $25.8 \pm 10.3$ & $4.0 \pm 1.3$ & $3.5 \pm 1.2$ \\
Mutagenicity & $28.5 \pm 14.1$ & $5.5 \pm 2.2$ & $2.9 \pm 1.1$ \\
NCI1 & $29.3 \pm 13.4$ & $4.3 \pm 1.5$ & $3.7 \pm 1.3$ \\
NCI-H23 & $25.8 \pm 10.3$ & $4.0 \pm 1.3$ & $3.5 \pm 1.2$ \\
OVCAR-8 & $25.8 \pm 10.3$ & $4.0 \pm 1.3$ & $3.5 \pm 1.2$ \\
PC-3 & $26.1 \pm 10.6$ & $4.1 \pm 1.4$ & $3.5 \pm 1.2$ \\
QM9 & $18.0 \pm 2.9$ & $5.3 \pm 1.5$ & $2.0 \pm 0.4$ \\
SF-295 & $25.8 \pm 10.3$ & $4.0 \pm 1.3$ & $3.5 \pm 1.2$ \\
SN12C & $25.8 \pm 10.3$ & $4.0 \pm 1.3$ & $3.5 \pm 1.2$ \\
Tox21 & $16.8 \pm 10.1$ & $3.0 \pm 1.2$ & $3.0 \pm 1.3$ \\
UACC257 & $25.8 \pm 10.3$ & $4.0 \pm 1.3$ & $3.5 \pm 1.2$ \\
Yeast & $21.1 \pm 8.8$ & $3.6 \pm 1.2$ & $3.3 \pm 1.1$ \\
\midrule
Peptides-func & $150.9 \pm 84.5$ & $5.7 \pm 2.6$ & $13.5 \pm 6.5$ \\
DD & $277.7 \pm 217.3$ & $36.0 \pm 20.7$ & $4.1 \pm 1.4$ \\
ENZYMES & $31.7 \pm 13.3$ & $5.4 \pm 2.2$ & $3.3 \pm 1.2$ \\
\midrule
KKI & $27.0 \pm 19.5$ & $7.2 \pm 5.1$ & $2.2 \pm 0.6$ \\
OHSU & $82.0 \pm 43.7$ & $20.0 \pm 13.2$ & $2.4 \pm 0.7$ \\
\bottomrule
\end{tabular}
}

\end{small}
\end{center}
\vskip -0.1in
\end{table}

To the best of our knowledge, the concept of graph bandwidth has been used in the context of GNNs only by \citet{balog2019fast}, with the explicit purpose of improving dense implementations on custom hardware. Their work does not leverage the bandwidth in the model itself and does not target graph generation.

\section{Bandwidth-Restricted Graph Generation} \label{sec:method}

In this section we describe \shortmethodname, our novel approach for improving graph generation. As \shortmethodname can be combined with different existing generative methods, we first describe its general strategy and principles in Section \ref{ssec:method-general}. Then, we detail the strategies for bandwidth-restricted graph generation applied to an autoregressive model based on GraphRNN \cite{pmlr-v80-you18a} in Section \ref{ssec:method-graphrnn}, and two distinct one-shot models based on Graphite \cite{pmlr-v97-grover19a} and EDP-GNN \cite{pmlr-v108-niu20a} in Section \ref{ssec:method-one-shot}.

\subsection{Restricting Graph Bandwidth} \label{ssec:method-general}
Starting from the observation that many real-world graphs have low bandwidth (Section \ref{sec:bandwidth}), we propose to reduce the output space of a graph generative model from $N \times N$ to $N \times \cmbandwidth$\footnote{Technically, as the adjacency matrix is symmetric, only a triangular matrix is generated both in the standard formulation and in our bandwidth-restricted re-parameterization.}. As the goal of graph generation is to learn a distribution of the data $p\left(G\right)$, we assume that, given $\cmbandwidth_{\text{data}}$ the maximum empirical bandwidth on the training set $\mathcal{G}_{\text{train}}$, we can reduce the output space of the generative model to $N \times \cmbandwidth_{\text{data}}$ without losing expressiveness (i.e., without losing the ability to generate in-distribution graphs). In general, we achieve this reduction through two complementary mechanisms: (1)~imposing a bandwidth-reducing ordering and (2)~restricting the output space to a dataset-specific bandwidth $\cmbandwidth_{\text{data}}$ or a graph-specific bandwidth $\cmbandwidth$ (Figure~\ref{fig:overview},~bottom).

During training, for each graph we use the precomputed adjacency matrix $A^{\pi^{*}}$ with  the ordering $\pi^{*}$ computed through the previously introduced C-M algorithm.
We remark that, given the linear time complexity of the C-M algorithm, our preprocessing does not introduce any additional overhead compared to using standard orderings such as BFS/DFS.  Notably, just restricting training examples to a specific canonical ordering does not guarantee that such an ordering will be respected during generation \cite{pmlr-v139-chen21j}, and does not provide any advantage in time complexity or output space reduction, since the complete adjacency matrix needs to be generated. Therefore, we also re-parameterize the adjacency matrix as $A^{\pi^{*}}_{\text{opt}} \in N \times \cmbandwidth_{\text{data}}$ (or $N \times \cmbandwidth$ for each graph), dropping the zeros outside the bandwidth. During sampling, only edges belonging to the reduced matrix are considered as candidates, thus constraining the output space and reducing the time complexity, without losing expressiveness.

Below, we discuss the details of our strategy applied to different models and highlight model-specific choices and advantages.

\subsection{Autoregressive Graph Generation} \label{ssec:method-graphrnn}

Autoregressive graph generation approaches recursively generate the edges of a single node \citep{pmlr-v80-you18a} or a group of nodes, \citep{liao2019efficient} conditioned on the previously generated subgraph.
This can be viewed as generating the adjacency matrix row-by-row or block-by-block.
We will focus on \mbox{GraphRNN} \citep{pmlr-v80-you18a}, although a similar approach could be applied to virtually any autoregressive graph generative model.

In \mbox{GraphRNN}, the probability of node $v_i$ being connected to node $v_j$, with $\pi(v_j) < \pi(v_i)$, is parameterized by an output function $f_\theta$ applied to the hidden state of an RNN over the previous rows of the adjacency matrix:
\[
    p[(v_i, v_j) \in \mathcal{E}] = f_\theta(\text{RNN}_{\phi}(A^\pi_{1:i -1}))_j,
\]
where $\theta$ and $\phi$ are parameters learned to maximize the likelihood of the data. In particular, we focus on the \mbox{GraphRNN-S} variant (additional model details are provided in Appendix~\ref{app:graphrnn-details}).

We note that a $d$-unit $f_\theta$ can generate graphs with at most bandwidth $d$.
Potentially, we could set $d$ to be the maximum number of nodes $N$ of any graph we want to generate, which would ensure maximum expressiveness.
Instead, we set $d$ equal to the maximum bandwidth of any $A^\pi$ we would want to generate, greatly increasing efficiency and reducing training signal sparsity for low bandwidth graphs.
We find the order $\pi$ for each graph $G$ by using the C-M algorithm and set \mbox{$d = \cmbandwidth_{\text{data}}$} as the maximum bandwidth across all  the $A^\pi$ in the training data.
Compared to generating $N$ rows of length $d = N$, we generate $\bigO(N / \cmbandwidth_\text{data})$-times fewer edges (corresponding to the savings factor, Table \ref{tab:naturalBWs}).

In the original GraphRNN model, \citet{pmlr-v80-you18a} used a random BFS ordering during training, and set \mbox{$d = M_{\text{data}} < N$}, with $M_{\text{data}}$ defined as the maximum number of nodes in the BFS queue at any time in the training data. 
$M_{\text{data}}$ is estimated empirically by sampling 100,000 BFS orderings per dataset and setting $M_{\text{data}}$ to be roughly the 99.9 percentile of the empirical distribution of maximum queue sizes. 
Critically, we observe that $M_{\text{data}}$ derived in \citet{pmlr-v80-you18a} approximates the \emph{maximum bandwidth across all possible BFS orderings}. In contrast, $\cmbandwidth_{\text{data}}$ is derived by explicitly reducing the bandwidth. Notably, our approach allows significantly shrinking $d$, thus directly reducing the output space and the time complexity. For example, on the DD dataset \cite{DOBSON2003771} of 918 protein graphs, the authors set $d = M_{\text{data}} = 230$, whereas we derive $d = \cmbandwidth_{\text{data}} = 122$, a nearly two-fold reduction. 

\subsection{One-shot Graph Generation} \label{ssec:method-one-shot}

One-shot graph generative models sample the entire graph topology simultaneously.
Typically the topology is represented by putting edge probabilities on each pair of nodes, resulting in a complete graph with a probability placed on each edge by the generative model \cite{graphVAE,pmlr-v97-grover19a,https://doi.org/10.48550/arxiv.1611.07308}.
We will call this graph the \emph{edge-probability graph}.
Our key insight is that the complete edge probability graph can be replaced with a bandwidth-restricted edge-probability graph (Figure~\ref{fig:overview}, bottom).

One-shot models can be divided into two main categories: 
(1)~\emph{Node-embedding-based} models sample node embeddings from the latent distribution and compute the edge-probability graph based on pairwise relationships, and 
(2)~\emph{Adjacency-matrix-based} models directly output the edge-probability graph. We focus on both categories, considering a model based on Graphite \citep{pmlr-v97-grover19a} and a model based on EDP-GNN \citep{pmlr-v108-niu20a}.

\textbf{Node-embedding-based one-shot generation.}\label{para:node-embed-one-shot}
Graphite \cite{pmlr-v97-grover19a} is a VAE-based approach with one latent vector $\mathbf{z_i} \in \mathbb{R}^d$ per node with standard Gaussian prior.
The encoder network uses graph convolution layers \citep{kipf2017semisupervised} on the input adjacency matrix $A \in \mathbb{R}^{N \times N}$ and node features $X \in \mathbb{R}^{N \times k}$ to derive the mean and standard deviation of the variational marginals:
\begin{equation}\label{eq:graphite-mu-sigma}
    \boldsymbol{\mu}, \boldsymbol{\sigma} = \text{GNN}_\phi(A,X),
\end{equation}
where $\boldsymbol{\mu}, \boldsymbol{\sigma} \in \mathbb{R}^{N \times d}$.
We will focus on the case without additional node features, so $X$ can be a positional encoding dependent on the node ordering.
The decoder network outputs edge probabilities:
\begin{equation}\label{eq:graphite_P_E}
    p[(v_i, v_j) \in \mathcal{E}] = \text{GNN}_\theta(\hat{A}, X, Z)_{i, j},
\end{equation}
where $\hat{A}$ is fully-connected and $Z$ are samples from $\mathcal{N}(\boldsymbol{\mu}, \boldsymbol{\sigma})$.
Further architectural details are described in Appendix \ref{app:graphite-details}.
In our bandwidth-restricted version of Graphite, we constrain the bandwidth of $\hat A$.
For a graph $G$, we get the bandwidth $\bandwidth(A^\pi)$ for $\pi$ found using the C-M algorithm and build a new edge set:
\begin{equation}
    \label{eq:bandwidth_edgeset}
    \hat{\mathcal{E}} = \{(i, j) \mid 1 \leq |i - j| \leq \bandwidth(A^\pi) \},
\end{equation}
where $\ 1 \leq i, j \leq N$.
With the new edge set, the decoder message passing steps are reduced from complexity $\bigO(N^2)$ to $\bigO(N\cdot \cmbandwidth)$.
During generation, the standard Graphite model samples the number of nodes from the empirical distribution of the training data. In our bandwidth-restricted version, both the number of nodes and the bandwidth $\bandwidth_{\text{data}}$ are sampled from the empirical distribution of the training data, thus further reducing the output space.

\textbf{Adjacency-matrix-based one-shot generation.} 
EDP-GNN \cite{pmlr-v108-niu20a} is a score-based model in which a GNN is trained to match the score function of the data distribution.
Intuitively, EDP-GNN learns to denoise the upper-right triangle of the adjacency matrix.
Our modification denoises the bandwidth-restricted upper-right triangle, thus reducing the modeled output space (Figure  \ref{fig:cosine-schedule}, Appendix).
In EDP-GNN, a multi-layer perceptron (MLP) predicts the final edge features from intermediate edge features $\hat A$ built by message passing layers and edge-update layers:
\begin{equation}
    \mathbf{s}_{\theta}(A)_{i, j} = \text{MLP}(\hat{A}_{ij}).
\end{equation}
To constrain the bandwidth, we restrict $(i, j) \in \hat{\mathcal{E}}$ (Eq. \ref{eq:bandwidth_edgeset}) in the final MLP and in all of the message passing and edge-update layers using the same approach described for Graphite. This drastically reduces the time complexity and output space.
Further details are included in Appendix \ref{app:edpgnn-details}.

\section{Experiments} \label{sec:experiments}

We experimentally validate our method on both synthetic and real graphs, comparing bandwidth-constrained architectures and non-constrained baselines.

\subsection{Metrics}
We measure two goals of the models: closeness of the sample distribution to the true distribution of graphs, and reconstruction quality.

To measure the quality of the sample distribution, we use two metrics. First, consistently with the recent literature, we use the Maximum Mean Discrepancy (MMD) between sampled and test graph statistics \cite{pmlr-v80-you18a}.
The graph statistics we compare are degree, clustering coefficient, node orbit counts following \citet{pmlr-v80-you18a}, and spectrum following \citet{liao2019efficient}.
To track overall sample quality, we compute the mean MMD$^2$ across all four MMD metrics\footnote{We note that previous works usually report MMD$^2$ but they indicate MMD in the results.}.
Additionally, as recommended by recent work on evaluation metrics for graph generative modeling \cite{thompson2022on}, we use a precision--recall metric \cite{NEURIPS2019_0234c510} which accounts for both sample quality and variety. We report the harmonic mean of precision and recall as F1-PR.

Reconstruction quality is measured by comparing the reconstructed graph and the original test graph using the Area Under the Precision--Recall Curve (AUPRC). For GraphRNN, we compare the  reconstructed row and the original row; for Graphite, we compare the reconstructed graph and the original graph\footnote{Reconstruction quality is less easily definable for 
 score-based models. Therefore, we omit AUPRC in EDP-GNN evaluation.}.
We chose to use AUPRC because it does not depend on true negatives and because it is well suited to class imbalance \cite{davis_relationship_2006}. Additionally, we report the estimated log-likelihood for test graphs. Although log-likelihood has known limitations for evaluating (graph) generative models \cite{thompson2022on}, it is well suited (in combination with the other metrics) to demonstrating the benefits of \shortmethodname’s reduced output space. 

\begin{table*}[t]  

\caption{Graph generation results on generic datasets. \textbf{Bold} indicates best results compared to the other model of the same type and dataset. Significance was determined by Welch's t-test with five replicates per model. Models are considered comparable when $p \ge 0.05$. MMD$^{2}$ denotes average MMD$^{2}$ across four metrics (degree, cluster, orbit, spectra). Due to space limitations, we provide all the individual metrics in Table \ref{tab:results-generic-extended} (Appendix).  OOM denotes out-of-memory issues. Hyphen (–) denotes not applicable metric/model.}
  
\label{tab:results-generic}
\vskip 0.1in
\begin{center}
\begin{small}

\resizebox{\textwidth}{!}{
\renewcommand{\arraystretch}{1.1}
\begin{tabular}{llcccccccccccc}
\toprule
 &  & \multicolumn{4}{c}{\textsc{community2}} & \multicolumn{4}{c}{\textsc{planar}} & \multicolumn{4}{c}{\textsc{grid2d}} \\
\cmidrule(l{2pt}r{2pt}){3-6}
\cmidrule(l{2pt}r{2pt}){7-10}
\cmidrule(l{2pt}r{2pt}){11-14}
 &  & $\downarrow$ MMD$^{2}$ & $\uparrow$ F1-PR & $\uparrow$ $\ell\ell$ & $\uparrow$ AUPRC & $\downarrow$ MMD$^{2}$ & $\uparrow$ F1-PR & $\uparrow$ $\ell\ell$ & $\uparrow$ AUPRC & $\downarrow$ MMD$^{2}$ & $\uparrow$ F1-PR & $\uparrow$ $\ell\ell$ & $\uparrow$ AUPRC \\
\midrule
\multirow{2}{*}{GraphRNN} & Standard & \textbf{0.028} & \textbf{0.648} & -1990 & 0.376 & 0.265 & \textbf{0.029} & -400.0 & 0.545 & 0.323 & 0.250 & -540.0 & 0.642 \\
 & \shortmethodname [ours] & \textbf{0.024} & \textbf{0.729} & \textbf{-1940} & \textbf{0.421} & \textbf{0.233} & \textbf{0.139} & \textbf{-309.0} & \textbf{0.652} & \textbf{0.240} & \textbf{0.909} & \textbf{-36.4} & \textbf{0.999} \\
\midrule
\multirow{2}{*}{Graphite} & Standard & 0.055 & \textbf{0.507} & -3560 & 0.706 & \textbf{0.361} & \textbf{0.010} & \textbf{-595.0} & 0.975 & 0.649 & 0.051 & -2320 & 0.453 \\
 & \shortmethodname [ours] & \textbf{0.047} & \textbf{0.423} & \textbf{-3450} & \textbf{0.747} & \textbf{0.468} & \textbf{0.023} & \textbf{-554.0} & \textbf{0.990} & \textbf{0.528} & \textbf{0.666} & \textbf{-358.0} & \textbf{0.915} \\
\midrule
\multirow{2}{*}{EDP-GNN} & Standard & \textbf{0.030} & \textbf{0.621} & -211000 & - & \textbf{0.459} & 0.172 & -57800 & - & \textbf{0.645} & \textbf{0.548} & -622000 & - \\
 & \shortmethodname [ours] & \textbf{0.040} & \textbf{0.581} & \textbf{-168000} & - & \textbf{0.474} & \textbf{0.450} & \textbf{-36400} & - & \textbf{0.590} & \textbf{0.528} & \textbf{-97900} & - \\
\midrule
 &  & \multicolumn{4}{c}{\textsc{DD}} & \multicolumn{4}{c}{\textsc{ENZYMES}} & \multicolumn{4}{c}{\textsc{PROTEINS}} \\
\cmidrule(l{2pt}r{2pt}){3-6}
\cmidrule(l{2pt}r{2pt}){7-10}
\cmidrule(l{2pt}r{2pt}){11-14}
 &  & $\downarrow$ MMD$^{2}$ & $\uparrow$ F1-PR & $\uparrow$ $\ell\ell$ & $\uparrow$ AUPRC & $\downarrow$ MMD$^{2}$ & $\uparrow$ F1-PR & $\uparrow$ $\ell\ell$ & $\uparrow$ AUPRC & $\downarrow$ MMD$^{2}$ & $\uparrow$ F1-PR & $\uparrow$ $\ell\ell$ & $\uparrow$ AUPRC \\
\midrule
\multirow{2}{*}{GraphRNN} & Standard & \textbf{0.174} & \textbf{0.426} & -1460 & \textbf{0.299} & 0.023 & \textbf{0.948} & -199.0 & 0.445 & \textbf{0.017} & \textbf{0.971} & -173.0 & 0.533 \\
 & \shortmethodname [ours] & \textbf{0.234} & \textbf{0.579} & \textbf{-1400} & \textbf{0.318} & \textbf{0.016} & \textbf{0.963} & \textbf{-177.0} & \textbf{0.602} & \textbf{0.020} & \textbf{0.964} & \textbf{-117.0} & \textbf{0.716} \\
\midrule
\multirow{2}{*}{Graphite} & Standard & \textbf{0.368} & \textbf{0.003} & -4360 & 0.804 & 0.107 & 0.459 & -204.0 & 0.925 & 0.153 & 0.523 & -220.0 & \textbf{0.950} \\
 & \shortmethodname [ours] & \textbf{0.273} & \textbf{0.008} & \textbf{-3430} & \textbf{0.840} & \textbf{0.038} & \textbf{0.916} & \textbf{-164.0} & \textbf{0.950} & \textbf{0.037} & \textbf{0.889} & \textbf{-166.0} & 0.933 \\
\midrule
\multirow{2}{*}{EDP-GNN} & Standard & OOM & OOM & OOM & OOM & \textbf{0.092} & 0.726 & -18000 & - & 0.077 & 0.782 & -23400 & - \\
 & \shortmethodname [ours] & \textbf{0.299} & \textbf{0.106} & \textbf{-269000} & - & \textbf{0.027} & \textbf{0.914} & \textbf{-7320} & - & \textbf{0.024} & \textbf{0.944} & \textbf{-7590} & - \\
\bottomrule
\end{tabular}
}

\end{small}
\end{center}
\end{table*}

\begin{table*}[ht]
  \caption{Graph generation results on molecular datasets. \textbf{Bold} indicates best results compared to the other model of the same type and dataset. Significance was determined by Welch's t-test with five replicates per model. Models are considered comparable when $p \ge 0.05$.
MMD$^{2}$ denotes average MMD$^{2}$ across four metrics (degree, cluster, orbit, spectra). Due to space limitations, we provide all the individual metrics in Table \ref{tab:results-generic-extended} (Appendix). Hyphen (–) denotes not applicable metric/model.}
  
\label{tab:results-molecular}
\vskip 0.1in
\begin{center}
\begin{small}

\resizebox{0.8\textwidth}{!}{

\renewcommand{\arraystretch}{1.1}
\begin{tabular}{llcccccccccccc}
\toprule
 &  & \multicolumn{4}{c}{\textsc{zinc250k}} & \multicolumn{4}{c}{\textsc{peptides-func}} \\
\cmidrule(l{2pt}r{2pt}){3-6}
\cmidrule(l{2pt}r{2pt}){7-10}
 &  & $\downarrow$ MMD$^{2}$ & $\uparrow$ F1-PR & $\uparrow$ $\ell\ell$ & $\uparrow$ AUPRC & $\downarrow$ MMD$^{2}$ & $\uparrow$ F1-PR & $\uparrow$ $\ell\ell$ & $\uparrow$ AUPRC \\
\midrule
\multirow{2}{*}{GraphRNN} & Standard & \textbf{0.038} & \textbf{0.803} & -38.9 & 0.668 & \textbf{0.031} & \textbf{0.477} & -168.0 & 0.809 \\
 & \shortmethodname [ours] & \textbf{0.029} & \textbf{0.900} & \textbf{-31.0} & \textbf{0.783} & \textbf{0.033} & \textbf{0.526} & \textbf{-112.0} & \textbf{0.903} \\
\midrule
\multirow{2}{*}{Graphite} & Standard & 0.153 & 0.178 & -57.0 & \textbf{0.999} & 0.182 & \textbf{0.031} & -742.0 & \textbf{0.987} \\
 & \shortmethodname [ours] & \textbf{0.084} & \textbf{0.511} & \textbf{-39.4} & 0.995 & \textbf{0.120} & \textbf{0.186} & \textbf{-362.0} & \textbf{0.993} \\
\midrule
\multirow{2}{*}{EDP-GNN} & Standard & \textbf{0.106} & 0.172 & -7510 & - & \textbf{0.115} & \textbf{0.041} & -634000 & - \\
 & \shortmethodname [ours] & \textbf{0.143} & \textbf{0.750} & \textbf{-3180} & - & \textbf{0.143} & \textbf{0.109} & \textbf{-32200} & - \\
\bottomrule
\end{tabular}
}

\end{small}
\end{center}
\end{table*}

\subsection{Experimental Setup}
We compare our bandwidth-restricted versions (+\shortmethodname) of the models based on GraphRNN \citep{pmlr-v80-you18a}, Graphite \citep{pmlr-v97-grover19a}, and EDP-GNN \citep{pmlr-v108-niu20a} to their standard baselines (i.e. without \shortmethodname). Our models are described in Section \ref{sec:method} and additional details are provided in Appendix \ref{app:model-details}.
Each model architecture is individually hyperoptimized (details in Appendix \ref{app:hyperopt}). All the experiments are repeated five times and significance is determined by Welch's $t$-test (models are considered comparable when $\text{p-value} \geq 0.05$).

\begin{figure*}[t]
\vskip 0.1in
\begin{center}
\centerline{\includegraphics[width=\textwidth]{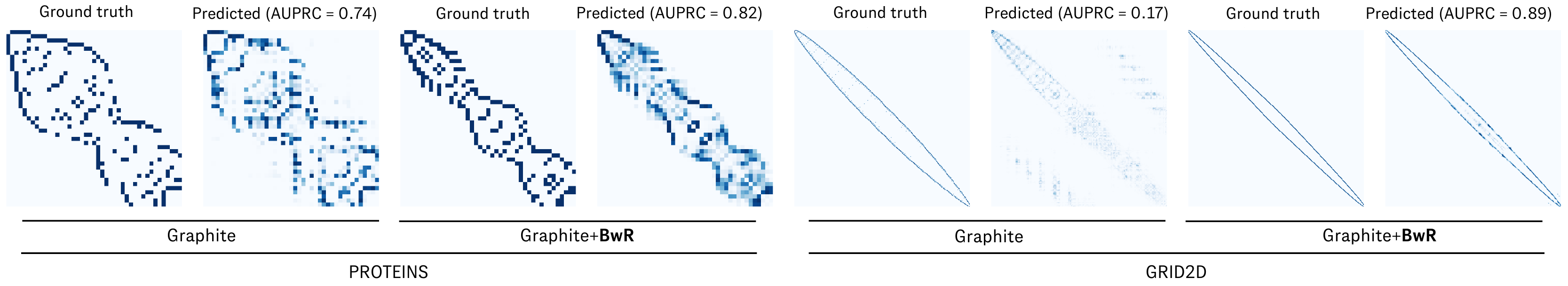}}
\caption{
    Comparison of Graphite reconstructions with and without \shortmethodname on samples from PROTEINS (left) and Grid2D (right).
}
\label{fig:recons}
\end{center}
\vskip -0.2in
\end{figure*}

\subsection{Generic Graph Generation} \label{ssec:results-generic}
\textbf{Datasets.} We evaluated our models on six standard  graph generation datasets, including both synthetic and real-world graphs: (1) \emph{Community2}, 1500 two-community graphs generated using an Erd\"os--Renyi model with  60--160 nodes; (2) \emph{Planar}, 1500 random planar graphs with 64 nodes made using Delaunay triangulation; (3) \emph{Grid2d}, 66 distinct two-dimensional grids with side lengths between 10 and 20; (4) \emph{DD}, 918 protein graphs with amino acids as nodes \cite{DOBSON2003771}; (5) \emph{Enzymes}, 556 protein graphs of enzyme tertiary structures from the BRENDA database \cite{schomburg2004brenda}; and (6) \emph{Proteins}, 904 protein graphs from the Protein Data Bank \cite{DOBSON2003771}. Additional details on the datasets are provided in Appendix \ref{app:generic-datasets}.

\textbf{Results.} Table \ref{tab:results-generic} summarizes the results for generic graph generation. As shown, our approach consistently achieves superior or competitive performance across datasets and methods as measured by mean MMD, F1-PR and/or AUPRC. GraphRNN+\shortmethodname and Graphite+\shortmethodname outperform their standard counterparts in five out of six datasets, with comparable performance  (i.e., not statistically significant or mixed) on the sixth. \shortmethodname improves EDP-GNN generation quality in four out of six datasets, with comparable performance on the others.  Notably, the dataset with the largest graphs, DD (mean 277.7 nodes per graph), could not be trained using standard \mbox{EDP-GNN} due to out-of-memory issues. In contrast, we are able to train our EDP-GNN+\shortmethodname, thus highlighting its lower memory complexity.
We observe a statistically significant improvement in AUPRC given by \shortmethodname in almost all the experiments, thus showing how our approach improves the accuracy of the reconstructed graphs even when bulk statistical distributions are comparable. Still, \shortmethodname results in an improvement in the quality of the sample distribution (MMD and/or F1-PR) in the majority of the experiments, with comparable performance in the others. We observe a significant improvement in log-likelihood compared to standard models across all the experiments.

We show examples of reconstructed adjacency matrices in Figure \ref{fig:recons}.
In the PROTEINS example (left), the standard model must predict edges in a much larger and imbalanced output space. In the Grid2d example (right), the standard model predicts edges far outside the bandwidth, which is inherently impossible with \shortmethodname.
Overall, results show that \shortmethodname consistently improves or matches generation quality at a fraction of the time/memory cost.

\subsection{Molecular Generation} \label{ssec:results-molecular}
\textbf{Datasets.} We evaluate our models on real-world molecular datasets to show the benefits of \shortmethodname for \textit{de~novo} molecular generation. We consider two datasets: \\
(1) \emph{ZINC250k} \cite{irwin2012zinc} includes 249,455 drug-like small molecules extracted from the ZINC database, averaging 23.14 heavy atoms (nodes) each. \\
(2) \emph{Peptides-func} \cite{dwivedi2022LRGB} is a recently published dataset of peptide structures that includes 15,535 molecules, averaging 150.94 heavy atoms (nodes) each. Compared to common small-molecule benchmarks, this dataset includes significantly larger molecular graphs and functional motifs (amino acids). Therefore, it allows us to test the advantages of a reduced time complexity and output space given by our bandwidth-constrained generation. \\
Additional datasets detailes are provided in Appendix~\ref{app:mol-datasets}.

\textbf{Results.} Table \ref{tab:results-molecular} shows the results on molecular graph generation. \shortmethodname achieves superior or competitive performance on both datasets across all methods. GraphRNN+\shortmethodname significantly improves   reconstruction accuracy with a comparable generation quality with respect to the standard baseline. Graphite+\shortmethodname leads to a significantly better generation quality with comparable or better reconstruction accuracy with respect to the standard baseline. Finally, EDP-GNN+\shortmethodname, achieves significantly improved generation quality on ZINC250k, and comparable results on Peptides-func.  We observe a significant improvement in log-likelihood compared to standard models across all the experiments.
We remark that, even when generation quality is comparable, our approach still significantly reduces time/memory complexity.

\begin{table*}[t]  

\caption{Computational cost results. \textbf{Bold} indicates best results compared to the other model of the same type and dataset. Significance was determined by Welch's t-test with five replicates per model. Models are considered comparable when $p \ge 0.05$. 
OOM denotes out-of-memory issues.}
  
\label{tab:results-computational}
\vskip 0.1in
\begin{center}
\begin{small}

\resizebox{0.9\textwidth}{!}{
\renewcommand{\arraystretch}{1.1}
\begin{tabular}{llcccccccc}
\toprule
 &  & \multicolumn{2}{c}{\textsc{community2}} & \multicolumn{2}{c}{\textsc{peptides-func}} & \multicolumn{2}{c}{\textsc{grid2d}} & \multicolumn{2}{c}{\textsc{DD}} \\
\cmidrule(l{2pt}r{2pt}){3-4}
\cmidrule(l{2pt}r{2pt}){5-6}
\cmidrule(l{2pt}r{2pt}){7-8}
\cmidrule(l{2pt}r{2pt}){9-10}
 &  & $\downarrow$ sample (s) & $\downarrow$ mem. (GB) & $\downarrow$ sample (s) & $\downarrow$ mem. (GB) & $\downarrow$ sample (s) & $\downarrow$ mem. (GB) & $\downarrow$ sample (s) & $\downarrow$ mem. (GB) \\
\midrule
\multirow{2}{*}{GraphRNN} & Standard & \textbf{0.673} & \textbf{0.059} & \textbf{7.690} & \textbf{0.082} & \textbf{0.490} & \textbf{0.115} & \textbf{1.470} & \textbf{0.142} \\
 & \shortmethodname [ours] & \textbf{0.538} & \textbf{0.061} & \textbf{7.680} & \textbf{0.080} & \textbf{0.520} & \textbf{0.112} & \textbf{1.500} & \textbf{0.149} \\
\midrule
\multirow{2}{*}{Graphite} & Standard & 1.460 & 0.805 & 3.520 & 1.740 & 6.400 & 3.270 & 10.90 & 5.870 \\
 & \shortmethodname [ours] & \textbf{1.050} & \textbf{0.553} & \textbf{0.268} & \textbf{0.170} & \textbf{0.842} & \textbf{0.447} & \textbf{2.380} & \textbf{1.510} \\
\midrule
\multirow{2}{*}{EDP-GNN} & Standard & 12.50 & 1.600 & 69.80 & 6.280 & 50.80 & 6.730 & OOM & OOM \\
 & \shortmethodname [ours] & \textbf{8.410} & \textbf{1.080} & \textbf{4.880} & \textbf{0.547} & \textbf{6.550} & \textbf{0.837} & \textbf{19.90} & \textbf{2.620} \\
\bottomrule
\end{tabular}
}

\end{small}
\end{center}
\end{table*}

\subsection{Memory and Time Efficiency}

We evaluate whether the lower time complexity and output space reduction actually translate into reduced time and memory footprint.
For this analysis, we consider all the datasets and models previously introduced, for a total of 24 experiments.
We measure the following metrics: (1)~\emph{Memory usage}, as the maximum GPU utilization during a training batch; (2)~\emph{Training time}, as the average wall time to train on one batch; (3)~\emph{Sample time}, as the average wall time to sample 256 graphs from the generative model.

Table~\ref{tab:results-computational} shows memory usage and sample time for four datasets. The remaining results are included in \cref{tab:results-computational_extended} (Appendix). As shown, \shortmethodname significantly reduces memory usage across all datasets and all methods besides GraphRNN\footnote{We believe that GraphRNN has close to the same time/memory usage with and without \shortmethodname because most of the computation happens in the hidden layers of the GRU as opposed to the single linear readout layer which predicts the next row.}, up to a factor of 11x for larger graphs. Additionally, it improves sample time (up to a factor of 13x-14x) in 14 out of 24 experiments (with comparable results in the others) and training time in 8 out of 24 experiments (with comparable results in the others). Overall, \shortmethodname achieves a consistent reduction in computational costs.

\subsection{Impact of Output Space Reduction to Performance }

We analyze the relationship between the savings factor, which summarizes the space reduction given by the C-M bandwidth reparameterization (Table \ref{tab:naturalBWs}), and the performance/computational  improvement. Results are included in Appendix~\ref{app:space-improvements}.
Results suggest that we can estimate the empirical improvement given by BwR for a specific dataset in advance, without actually training a model, by computing simple statistical properties of the dataset.

\section{Conclusion} \label{sec:conclusion}
We presented \shortmethodname, a novel approach to reduce the time/space complexity and output space of graph generative models.  Leveraging the observation that many real-world graphs have low graph bandwidth, our method restricts the bandwidth of the adjacency matrices during training and generation. Our method is compatible with virtually all existing graph generative models, and we described its application to autoregressive, VAE-based, and score-based models. Our extensive results on both synthetic, biological, and chemical datasets showed that our strategy consistently achieves superior or comparable generation quality compared to the standard methods, while significantly reducing time/space complexity. Currently, our strategy leverages random Cuthill-McKee orderings to reduce the bandwidth. Future work will explore other---potentially even learned---bandwidth-minimizing orderings, while further theoretically studying the distribution of orderings induced by our approach. Future work will also extend our strategy to additional settings, such as conditional generation, larger graphs, and new state-of-the-art models. 

\section*{Acknowledgements}



We thank the anonymous reviewers for their suggestions which have further strengthened the conclusions of the paper. We thank members of the AI/ML department in Genentech for helpful feedback and discussions. All authors are employees of Genentech, Inc. and shareholders of Roche.

\FloatBarrier
\bibliography{main}

\begin{thebibliography}{57}
\providecommand{\natexlab}[1]{#1}
\providecommand{\url}[1]{\texttt{#1}}
\expandafter\ifx\csname urlstyle\endcsname\relax
  \providecommand{\doi}[1]{doi: #1}\else
  \providecommand{\doi}{doi: \begingroup \urlstyle{rm}\Url}\fi

\bibitem[Albert \& Barab{\'a}si(2002)Albert and
  Barab{\'a}si]{albert2002statistical}
Albert, R. and Barab{\'a}si, A.-L.
\newblock Statistical mechanics of complex networks.
\newblock \emph{Reviews of modern physics}, 74\penalty0 (1):\penalty0 47, 2002.

\bibitem[Balog et~al.(2019)Balog, van Merri{\"e}nboer, Moitra, Li, and
  Tarlow]{balog2019fast}
Balog, M., van Merri{\"e}nboer, B., Moitra, S., Li, Y., and Tarlow, D.
\newblock Fast training of sparse graph neural networks on dense hardware.
\newblock \emph{arXiv preprint arXiv:1906.11786}, 2019.

\bibitem[Biewald(2020)]{wandb}
Biewald, L.
\newblock Experiment tracking with weights and biases, 2020.
\newblock URL \url{https://www.wandb.com/}.
\newblock Software available from wandb.com.

\bibitem[B{\"o}ttcher et~al.(2010)B{\"o}ttcher, Pruessmann, Taraz, and
  W{\"u}rfl]{bottcher2010bandwidth}
B{\"o}ttcher, J., Pruessmann, K.~P., Taraz, A., and W{\"u}rfl, A.
\newblock Bandwidth, expansion, treewidth, separators and universality for
  bounded-degree graphs.
\newblock \emph{European Journal of Combinatorics}, 31\penalty0 (5):\penalty0
  1217--1227, 2010.

\bibitem[Chan \& George(1980)Chan and George]{chan1980linear}
Chan, W.-M. and George, A.
\newblock {A linear time implementation of the reverse Cuthill-McKee
  algorithm}.
\newblock \emph{BIT Numerical Mathematics}, 20\penalty0 (1):\penalty0 8--14,
  1980.

\bibitem[Chen et~al.(2021)Chen, Han, Hu, Ruiz, and Liu]{pmlr-v139-chen21j}
Chen, X., Han, X., Hu, J., Ruiz, F., and Liu, L.
\newblock Order matters: probabilistic modeling of node sequence for graph
  generation.
\newblock In \emph{Proceedings of the 38th International Conference on Machine
  Learning}, volume 139 of \emph{Proceedings of Machine Learning Research},
  pp.\  1630--1639. PMLR, 2021.

\bibitem[Cuthill \& McKee(1969)Cuthill and McKee]{cuthill1969reducing}
Cuthill, E. and McKee, J.
\newblock Reducing the bandwidth of sparse symmetric matrices.
\newblock In \emph{Proceedings of the 1969 24th national conference}, pp.\
  157--172, 1969.

\bibitem[Cygan \& Pilipczuk(2010)Cygan and Pilipczuk]{cygan2010exact}
Cygan, M. and Pilipczuk, M.
\newblock Exact and approximate bandwidth.
\newblock \emph{Theoretical Computer Science}, 411\penalty0 (40-42):\penalty0
  3701--3713, 2010.

\bibitem[Dai et~al.(2020)Dai, Nazi, Li, Dai, and Schuurmans]{dai2020scalable}
Dai, H., Nazi, A., Li, Y., Dai, B., and Schuurmans, D.
\newblock Scalable deep generative modeling for sparse graphs.
\newblock In \emph{Proceedings of the 37th International Conference on Machine
  Learning}, volume 119 of \emph{Proceedings of Machine Learning Research},
  pp.\  2302--2312. PMLR, 2020.

\bibitem[Davis \& Goadrich(2006)Davis and Goadrich]{davis_relationship_2006}
Davis, J. and Goadrich, M.
\newblock The relationship between {Precision}-{Recall} and {ROC} curves.
\newblock In \emph{{International Conference on Machine Learning}}, pp.\
  233--240. Association for Computing Machinery, 2006.

\bibitem[Dobson \& Doig(2003)Dobson and Doig]{DOBSON2003771}
Dobson, P.~D. and Doig, A.~J.
\newblock Distinguishing enzyme structures from non-enzymes without alignments.
\newblock \emph{Journal of Molecular Biology}, 330\penalty0 (4):\penalty0
  771--783, 2003.

\bibitem[Dwivedi et~al.(2022)Dwivedi, Ramp\'{a}\v{s}ek, Galkin, Parviz, Wolf,
  Luu, and Beaini]{dwivedi2022LRGB}
Dwivedi, V.~P., Ramp\'{a}\v{s}ek, L., Galkin, M., Parviz, A., Wolf, G., Luu,
  A.~T., and Beaini, D.
\newblock Long range graph benchmark.
\newblock In \emph{Advances in Neural Information Processing Systems},
  volume~35, pp.\  22326--22340. Curran Associates, Inc., 2022.

\bibitem[Eppstein(2009)]{eppstein2009isometric}
Eppstein, D.
\newblock Isometric diamond subgraphs.
\newblock In \emph{Graph Drawing: 16th International Symposium, GD 2008,
  Heraklion, Crete, Greece, September 21-24, 2008. Revised Papers 16}, pp.\
  384--389. Springer, 2009.

\bibitem[Feige(2000)]{feige2000coping}
Feige, U.
\newblock Coping with the {NP-hardness} of the graph bandwidth problem.
\newblock In \emph{Scandinavian Workshop on Algorithm Theory}, pp.\  10--19.
  Springer, 2000.

\bibitem[Fey \& Lenssen(2019)Fey and Lenssen]{Fey/Lenssen/2019}
Fey, M. and Lenssen, J.~E.
\newblock Fast graph representation learning with {PyTorch Geometric}.
\newblock In \emph{ICLR Workshop on Representation Learning on Graphs and
  Manifolds}, 2019.

\bibitem[Gibbs et~al.(1976)Gibbs, Poole, and Stockmeyer]{10.2307/2156090}
Gibbs, N.~E., Poole, W.~G., and Stockmeyer, P.~K.
\newblock An algorithm for reducing the bandwidth and profile of a sparse
  matrix.
\newblock \emph{SIAM Journal on Numerical Analysis}, 13\penalty0 (2):\penalty0
  236--250, 1976.

\bibitem[G{\'o}mez-Bombarelli et~al.(2018)G{\'o}mez-Bombarelli, Wei, Duvenaud,
  Hern{\'a}ndez-Lobato, S{\'a}nchez-Lengeling, Sheberla, Aguilera-Iparraguirre,
  Hirzel, Adams, and Aspuru-Guzik]{gomez2018automatic}
G{\'o}mez-Bombarelli, R., Wei, J.~N., Duvenaud, D., Hern{\'a}ndez-Lobato,
  J.~M., S{\'a}nchez-Lengeling, B., Sheberla, D., Aguilera-Iparraguirre, J.,
  Hirzel, T.~D., Adams, R.~P., and Aspuru-Guzik, A.
\newblock Automatic chemical design using a data-driven continuous
  representation of molecules.
\newblock \emph{ACS central science}, 4\penalty0 (2):\penalty0 268--276, 2018.

\bibitem[Gonzaga~de Oliveira et~al.(2018)Gonzaga~de Oliveira, Bernardes, and
  Chagas]{gonzaga2018evaluation}
Gonzaga~de Oliveira, S.~L., Bernardes, J.~A., and Chagas, G.~O.
\newblock An evaluation of low-cost heuristics for matrix bandwidth and profile
  reductions.
\newblock \emph{Computational and Applied Mathematics}, 37\penalty0
  (2):\penalty0 1412--1471, 2018.

\bibitem[Goyal et~al.(2020)Goyal, Jain, and Ranu]{goyal2020graphgen}
Goyal, N., Jain, H.~V., and Ranu, S.
\newblock {GraphGen}: a scalable approach to domain-agnostic labeled graph
  generation.
\newblock In \emph{Proceedings of The Web Conference 2020}, pp.\  1253--1263,
  2020.

\bibitem[Grover et~al.(2019)Grover, Zweig, and Ermon]{pmlr-v97-grover19a}
Grover, A., Zweig, A., and Ermon, S.
\newblock Graphite: iterative generative modeling of graphs.
\newblock In \emph{Proceedings of the 36th International Conference on Machine
  Learning}, volume~97 of \emph{Proceedings of Machine Learning Research}, pp.\
   2434--2444. PMLR, 2019.

\bibitem[Guo \& Zhao(2023)Guo and Zhao]{guo2022systematic}
Guo, X. and Zhao, L.
\newblock A systematic survey on deep generative models for graph generation.
\newblock \emph{IEEE Transactions on Pattern Analysis and Machine
  Intelligence}, 45\penalty0 (5):\penalty0 5370--5390, 2023.

\bibitem[Hamilton et~al.(2017)Hamilton, Ying, and
  Leskovec]{DBLP:journals/debu/HamiltonYL17}
Hamilton, W.~L., Ying, R., and Leskovec, J.
\newblock Representation learning on graphs: Methods and applications.
\newblock \emph{{IEEE} Data Eng. Bull.}, 40\penalty0 (3):\penalty0 52--74,
  2017.

\bibitem[Hendrycks \& Gimpel(2016)Hendrycks and
  Gimpel]{hendrycks_gaussian_2020}
Hendrycks, D. and Gimpel, K.
\newblock Gaussian error linear units ({GELUs}).
\newblock \emph{arXiv preprint arXiv:1606.08415}, 2016.

\bibitem[Ho et~al.(2020)Ho, Jain, and Abbeel]{ddpm-ho-2020}
Ho, J., Jain, A., and Abbeel, P.
\newblock Denoising diffusion probabilistic models.
\newblock In \emph{Advances in Neural Information Processing Systems},
  volume~33, pp.\  6840--6851. Curran Associates, Inc., 2020.

\bibitem[Hu et~al.(2020)Hu, Liu, Gomes, Zitnik, Liang, Pande, and
  Leskovec]{Hu*2020Strategies}
Hu, W., Liu, B., Gomes, J., Zitnik, M., Liang, P., Pande, V., and Leskovec, J.
\newblock Strategies for pre-training graph neural networks.
\newblock In \emph{International Conference on Learning Representations}, 2020.

\bibitem[Irwin et~al.(2012)Irwin, Sterling, Mysinger, Bolstad, and
  Coleman]{irwin2012zinc}
Irwin, J.~J., Sterling, T., Mysinger, M.~M., Bolstad, E.~S., and Coleman, R.~G.
\newblock {ZINC}: a free tool to discover chemistry for biology.
\newblock \emph{Journal of Chemical Information and Modeling}, 52\penalty0
  (7):\penalty0 1757--1768, 2012.

\bibitem[Jia et~al.(2020)Jia, Lin, Gao, Zaharia, and Aiken]{jia2020improving}
Jia, Z., Lin, S., Gao, M., Zaharia, M., and Aiken, A.
\newblock Improving the accuracy, scalability, and performance of graph neural
  networks with {ROC}.
\newblock \emph{Proceedings of Machine Learning and Systems}, 2:\penalty0
  187--198, 2020.

\bibitem[Jin et~al.(2018)Jin, Barzilay, and Jaakkola]{jin2018junction}
Jin, W., Barzilay, R., and Jaakkola, T.
\newblock Junction tree variational autoencoder for molecular graph generation.
\newblock In \emph{Proceedings of the 35th International Conference on Machine
  Learning}, volume~80 of \emph{Proceedings of Machine Learning Research}, pp.\
   2323--2332. PMLR, 2018.

\bibitem[Kipf \& Welling(2016)Kipf and
  Welling]{https://doi.org/10.48550/arxiv.1611.07308}
Kipf, T.~N. and Welling, M.
\newblock Variational graph auto-encoders.
\newblock In \emph{Neural Information Processing Systems Workshop on Bayesian
  Deep Learning}, 2016.

\bibitem[Kipf \& Welling(2017)Kipf and Welling]{kipf2017semisupervised}
Kipf, T.~N. and Welling, M.
\newblock Semi-supervised classification with graph convolutional networks.
\newblock In \emph{International Conference on Learning Representations}, 2017.

\bibitem[Kynk\"{a}\"{a}nniemi et~al.(2019)Kynk\"{a}\"{a}nniemi, Karras, Laine,
  Lehtinen, and Aila]{NEURIPS2019_0234c510}
Kynk\"{a}\"{a}nniemi, T., Karras, T., Laine, S., Lehtinen, J., and Aila, T.
\newblock Improved precision and recall metric for assessing generative models.
\newblock In \emph{Advances in Neural Information Processing Systems},
  volume~32. Curran Associates, Inc., 2019.

\bibitem[Landrum(2006)]{greg-landrum-rdkit}
Landrum, G.
\newblock {RDKit}: Open-source cheminformatics, 2006.
\newblock URL \url{https://rdkit.org}.

\bibitem[Li et~al.(2022)Li, Huang, and Zitnik]{li2022graph}
Li, M.~M., Huang, K., and Zitnik, M.
\newblock Graph representation learning in biomedicine and healthcare.
\newblock \emph{Nature biomedical engineering}, 6\penalty0 (12):\penalty0
  1353--1369, 2022.

\bibitem[Li et~al.(2018)Li, Vinyals, Dyer, Pascanu, and
  Battaglia]{li2018learning}
Li, Y., Vinyals, O., Dyer, C., Pascanu, R., and Battaglia, P.
\newblock Learning deep generative models of graphs.
\newblock \emph{arXiv preprint arXiv:1803.03324}, 2018.

\bibitem[Liao et~al.(2019)Liao, Li, Song, Wang, Hamilton, Duvenaud, Urtasun,
  and Zemel]{liao2019efficient}
Liao, R., Li, Y., Song, Y., Wang, S., Hamilton, W., Duvenaud, D.~K., Urtasun,
  R., and Zemel, R.
\newblock Efficient graph generation with graph recurrent attention networks.
\newblock In \emph{Advances in Neural Information Processing Systems},
  volume~32. Curran Associates, Inc., 2019.

\bibitem[Liu et~al.(2018)Liu, Allamanis, Brockschmidt, and
  Gaunt]{liu2018constrained}
Liu, Q., Allamanis, M., Brockschmidt, M., and Gaunt, A.
\newblock Constrained graph variational autoencoders for molecule design.
\newblock In \emph{Advances in Neural Information Processing Systems},
  volume~31. Curran Associates, Inc., 2018.

\bibitem[Loshchilov \& Hutter(2017)Loshchilov and Hutter]{loshchilov2017sgdr}
Loshchilov, I. and Hutter, F.
\newblock {SGDR}: Stochastic gradient descent with warm restarts.
\newblock In \emph{International Conference on Learning Representations}, 2017.

\bibitem[Loshchilov \& Hutter(2019)Loshchilov and
  Hutter]{loshchilov2018decoupled}
Loshchilov, I. and Hutter, F.
\newblock Decoupled weight decay regularization.
\newblock In \emph{International Conference on Learning Representations}, 2019.

\bibitem[Ma et~al.(2018)Ma, Chen, and Xiao]{ma2018constrained}
Ma, T., Chen, J., and Xiao, C.
\newblock Constrained generation of semantically valid graphs via regularizing
  variational autoencoders.
\newblock In \emph{Advances in Neural Information Processing Systems},
  volume~31. Curran Associates, Inc., 2018.

\bibitem[Mercado et~al.(2021)Mercado, Bjerrum, and
  Engkvist]{mercado2021exploring}
Mercado, R., Bjerrum, E.~J., and Engkvist, O.
\newblock Exploring graph traversal algorithms in graph-based molecular
  generation.
\newblock \emph{Journal of Chemical Information and Modeling}, 62\penalty0
  (9):\penalty0 2093--2100, 2021.

\bibitem[Monien(1986)]{monien1986bandwidth}
Monien, B.
\newblock The bandwidth minimization problem for caterpillars with hair length
  3 is {NP}-complete.
\newblock \emph{SIAM Journal on Algebraic Discrete Methods}, 7\penalty0
  (4):\penalty0 505--512, 1986.

\bibitem[Morris et~al.(2020)Morris, Kriege, Bause, Kersting, Mutzel, and
  Neumann]{morris2020tudataset}
Morris, C., Kriege, N.~M., Bause, F., Kersting, K., Mutzel, P., and Neumann, M.
\newblock {TUDataset}: a collection of benchmark datasets for learning with
  graphs.
\newblock In \emph{ICML 2020 Workshop Graph Representation Learning and
  Beyond}, 2020.

\bibitem[Nichol \& Dhariwal(2021)Nichol and Dhariwal]{nichol2021improved}
Nichol, A.~Q. and Dhariwal, P.
\newblock Improved denoising diffusion probabilistic models.
\newblock In \emph{Proceedings of the 38th International Conference on Machine
  Learning}, volume 139 of \emph{Proceedings of Machine Learning Research},
  pp.\  8162--8171. PMLR, 2021.

\bibitem[Niu et~al.(2020)Niu, Song, Song, Zhao, Grover, and
  Ermon]{pmlr-v108-niu20a}
Niu, C., Song, Y., Song, J., Zhao, S., Grover, A., and Ermon, S.
\newblock Permutation invariant graph generation via score-based generative
  modeling.
\newblock In \emph{Proceedings of the International Conference on Artificial
  Intelligence and Statistics}, pp.\  4474--4484. PMLR, 2020.

\bibitem[Otachi \& Suda(2011)Otachi and Suda]{otachi2011bandwidth}
Otachi, Y. and Suda, R.
\newblock Bandwidth and pathwidth of three-dimensional grids.
\newblock \emph{Discrete Mathematics}, 311\penalty0 (10-11):\penalty0 881--887,
  2011.

\bibitem[Papadimitriou(1976)]{papadimitriou1976np}
Papadimitriou, C.~H.
\newblock The {NP}-completeness of the bandwidth minimization problem.
\newblock \emph{Computing}, 16\penalty0 (3):\penalty0 263--270, 1976.

\bibitem[Paszke et~al.(2019)Paszke, Gross, Massa, Lerer, Bradbury, Chanan,
  Killeen, Lin, Gimelshein, Antiga, Desmaison, Kopf, Yang, DeVito, Raison,
  Tejani, Chilamkurthy, Steiner, Fang, Bai, and
  Chintala]{Paszke_PyTorch_An_Imperative_2019}
Paszke, A., Gross, S., Massa, F., Lerer, A., Bradbury, J., Chanan, G., Killeen,
  T., Lin, Z., Gimelshein, N., Antiga, L., Desmaison, A., Kopf, A., Yang, E.,
  DeVito, Z., Raison, M., Tejani, A., Chilamkurthy, S., Steiner, B., Fang, L.,
  Bai, J., and Chintala, S.
\newblock {PyTorch}: an imperative style, high-performance deep learning
  library.
\newblock In Wallach, H., Larochelle, H., Beygelzimer, A., d\textquotesingle
  Alch\'{e}-Buc, F., Fox, E., and Garnett, R. (eds.), \emph{Advances in Neural
  Information Processing Systems}, volume~32. Curran Associates, Inc., 2019.

\bibitem[Schomburg et~al.(2004)Schomburg, Chang, Ebeling, Gremse, Heldt, Huhn,
  and Schomburg]{schomburg2004brenda}
Schomburg, I., Chang, A., Ebeling, C., Gremse, M., Heldt, C., Huhn, G., and
  Schomburg, D.
\newblock {BRENDA}, the enzyme database: updates and major new developments.
\newblock \emph{Nucleic acids research}, 32\penalty0 (suppl\_1):\penalty0
  D431--D433, 2004.

\bibitem[Shi et~al.(2020)Shi, Xu, Zhu, Zhang, Zhang, and
  Tang]{Shi*2020GraphAF:}
Shi, C., Xu, M., Zhu, Z., Zhang, W., Zhang, M., and Tang, J.
\newblock {GraphAF}: a flow-based autoregressive model for molecular graph
  generation.
\newblock In \emph{International Conference on Learning Representations}, 2020.

\bibitem[Simonovsky \& Komodakis(2018)Simonovsky and Komodakis]{graphVAE}
Simonovsky, M. and Komodakis, N.
\newblock Graphvae: Towards generation of small graphs using variational
  autoencoders.
\newblock In K{\r{u}}rkov{\'a}, V., Manolopoulos, Y., Hammer, B., Iliadis, L.,
  and Maglogiannis, I. (eds.), \emph{Artificial Neural Networks and Machine
  Learning -- ICANN 2018}, pp.\  412--422. Springer International Publishing,
  2018.

\bibitem[Thompson et~al.(2022)Thompson, Knyazev, Ghalebi, Kim, and
  Taylor]{thompson2022on}
Thompson, R., Knyazev, B., Ghalebi, E., Kim, J., and Taylor, G.~W.
\newblock On evaluation metrics for graph generative models.
\newblock In \emph{International Conference on Learning Representations}, 2022.

\bibitem[Turner(1986)]{turner1986probable}
Turner, J.~S.
\newblock On the probable performance of heuristics for bandwidth minimization.
\newblock \emph{SIAM journal on computing}, 15\penalty0 (2):\penalty0 561--580,
  1986.

\bibitem[Unger(1998)]{unger1998complexity}
Unger, W.
\newblock The complexity of the approximation of the bandwidth problem.
\newblock In \emph{Proceedings 39th Annual Symposium on Foundations of Computer
  Science (Cat. No. 98CB36280)}, pp.\  82--91. IEEE, 1998.

\bibitem[Vaswani et~al.(2017)Vaswani, Shazeer, Parmar, Uszkoreit, Jones, Gomez,
  Kaiser, and Polosukhin]{attention_vaswani}
Vaswani, A., Shazeer, N., Parmar, N., Uszkoreit, J., Jones, L., Gomez, A.~N.,
  Kaiser, L.~u., and Polosukhin, I.
\newblock Attention is all you need.
\newblock In \emph{Advances in Neural Information Processing Systems},
  volume~30. Curran Associates, Inc., 2017.

\bibitem[Wester et~al.(2008)Wester, Pollock, Coutsias, Allu, Muresan, and
  Oprea]{wester2008scaffold}
Wester, M.~J., Pollock, S.~N., Coutsias, E.~A., Allu, T.~K., Muresan, S., and
  Oprea, T.~I.
\newblock Scaffold topologies. 2. analysis of chemical databases.
\newblock \emph{Journal of Chemical Information and Modeling}, 48\penalty0
  (7):\penalty0 1311--1324, 2008.

\bibitem[Winter et~al.(2021)Winter, Noe, and Clevert]{winter2021permutation}
Winter, R., Noe, F., and Clevert, D.-A.
\newblock Permutation-invariant variational autoencoder for graph-level
  representation learning.
\newblock In \emph{Advances in Neural Information Processing Systems},
  volume~34, pp.\  9559--9573. Curran Associates, Inc., 2021.

\bibitem[You et~al.(2018)You, Ying, Ren, Hamilton, and
  Leskovec]{pmlr-v80-you18a}
You, J., Ying, R., Ren, X., Hamilton, W., and Leskovec, J.
\newblock {G}raph{RNN}: generating realistic graphs with deep auto-regressive
  models.
\newblock In \emph{International Conference on Machine Learning}, pp.\
  5708--5717. PMLR, 2018.

\end{thebibliography}
\bibliographystyle{icml2023}

\newpage
\appendix
\onecolumn

\section{Bandwidth Visualization}\label{app:bw-visualization}
\begin{figure}[h]
\vskip 0.2in
\begin{center}
\centerline{
\includegraphics[width=0.8\textwidth]{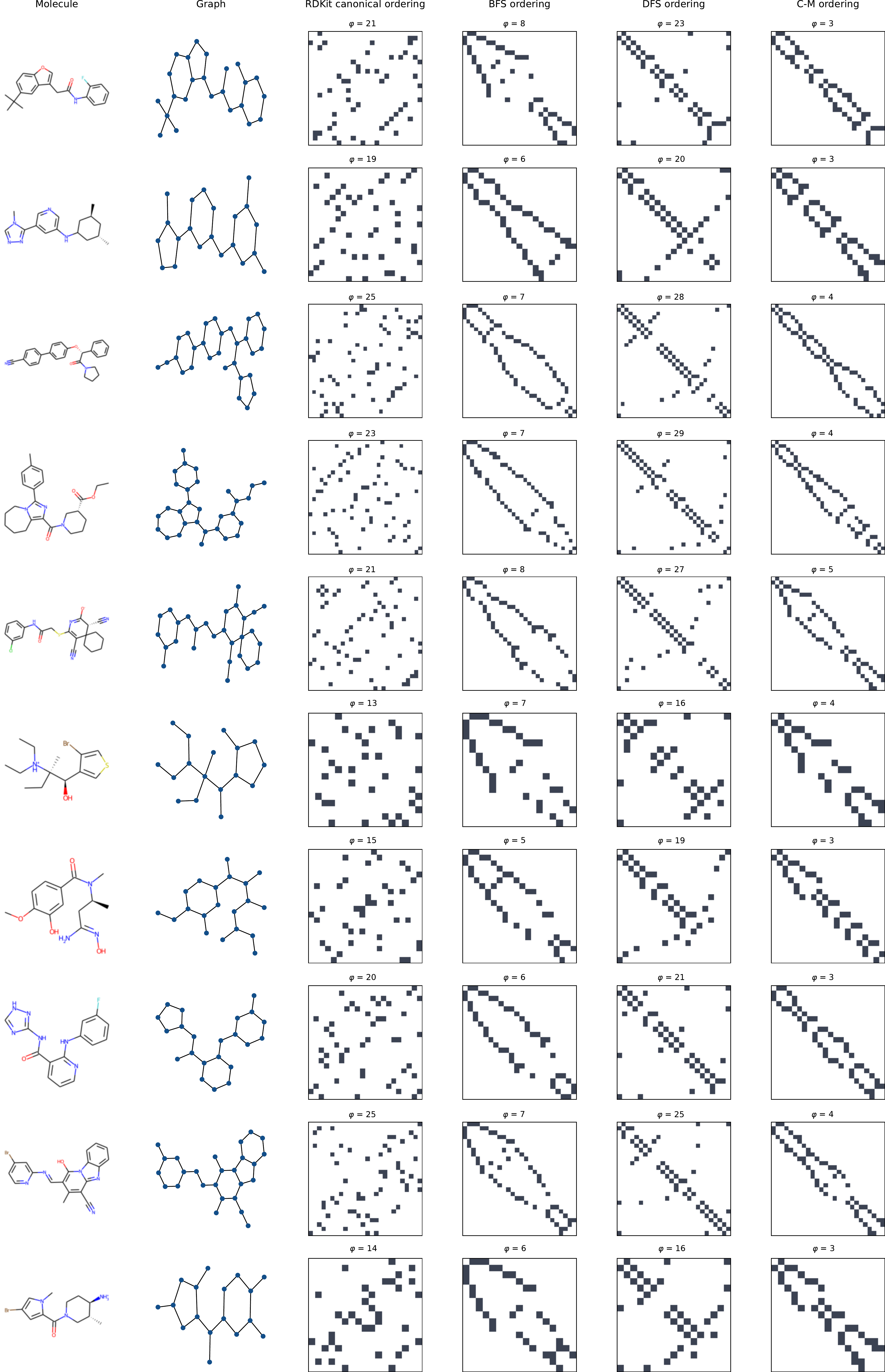}
}
\caption{Adjacency matrix and bandwidth $\bandwidth$ for different orderings (canonical RDKit, BFS, DFS and Cuthill-McKee) for a random set of 10 molecules from ZINC250k.}
\label{fig:mol-orderings}
\end{center}
\vskip -0.2in
\end{figure}

In Figure \ref{fig:mol-orderings}, we show the adjacency matrix $A$ and the bandwidth $\bandwidth(A)$ for a random set of 10 molecules from ZINC250k. The RDKit ordering is computed using the canonical atom ranking provided by the RDKit library. For the BFS, DFS and C-M, we randomly sample 100 orderings and plot the one with the highest $\bandwidth$ (this approximates the output space needed to correctly model the graph for each ordering).

\begin{figure}[h]
\vskip 0.1in
\begin{center}
\centerline{
\includegraphics[width=0.6\columnwidth]{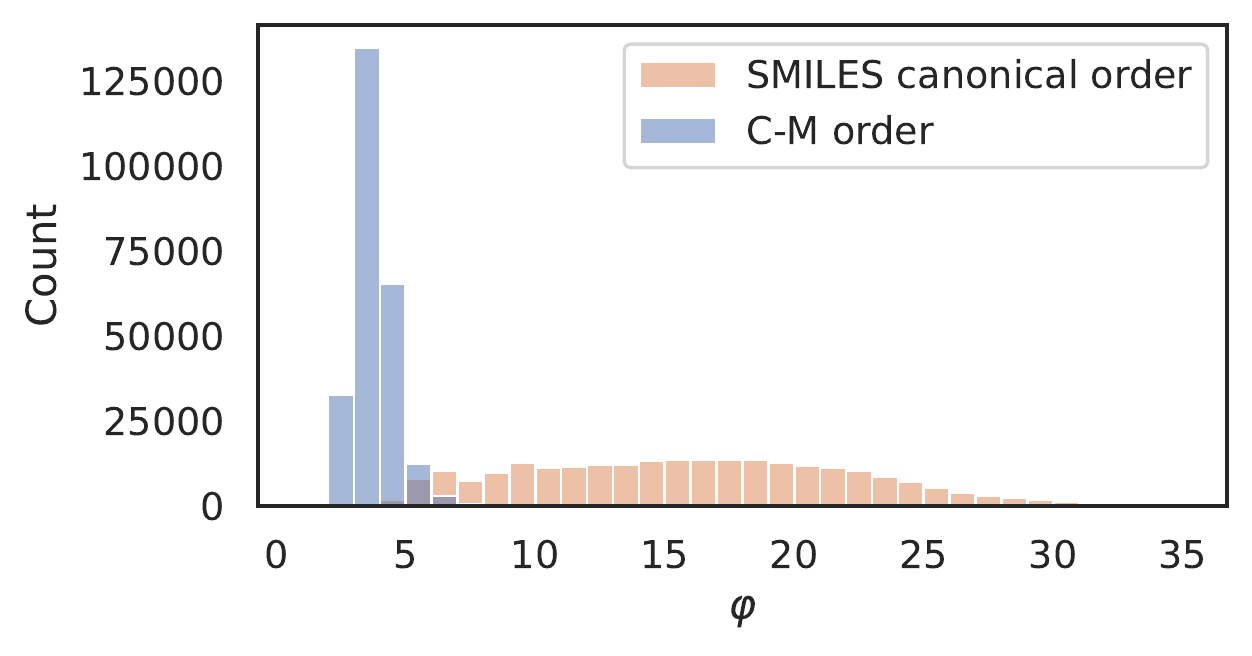}}
\caption{Distribution of bandwidth of ZINC250k adjacency matrices using the C-M algorithm and the canonical SMILES order.}
\label{fig:zincBW}
\end{center}
\vskip -0.2in
\end{figure}

In Figure \ref{fig:zincBW}, we show the distribution of bandwidth of ZINC250k adjacency matrices using the C-M algorithm and the canonical SMILES order.

\section{Model Details}\label{app:model-details}
Each model was re-implemented and somewhat modified to facilitate the pairwise comparison with and without the \shortmethodname modification.

\subsection{Optimization}
All models were trained for 100 epochs of 30 training batches and nine validation batches.
The batch size was fixed at 32.
The AdamW optimizer \citep{loshchilov2018decoupled} was used with a cosine annealed learning rate \citep{loshchilov2017sgdr}.
The initial learning rate was hyperoptimized (Appendix \ref{app:hyperopt}), and the weight decay parameter was either set to zero or hyperoptimized.
GraphRNN and Graphite were trained using binary cross entropy to measure reconstruction accuracy.
EDP-GNN was trained using mean squared error loss.

\subsection{Hyperoptimization}\label{app:hyperopt}

Hyperparameters were separately optimized for each combination of node order, model, and dataset.
The hyperparameters for MMD and AUPRC results were chosen to minimize mean validation MMD$^2$ in the case of GraphRNN and EDP-GNN, and MMD$^2$ $-$ AUPRC in the case of Graphite. The hyperparameters for log-likelihood and F1-PR results were chosen to maximize F1-PR. All hyperoptimizations used 20 outer-loop steps of the Weights and Biases \citep{wandb} Bayesian hyperoptimizer.
For the specific hyperparameter ranges, see the model details below.

\subsection{GraphRNN} \label{app:graphrnn-details}
GraphRNN \cite{pmlr-v80-you18a} is an autoregressive model for generating adjacency matrices.
We used the GraphRNN-S variant which uses an MLP to predict a whole row at once of the adjacency matrix from the RNN's hidden state.

\paragraph{GraphRNN data pre-processing.}
GraphRNN was trained using teacher forcing to autogressively predict the next row of the re-parameterized adjacency matrices $A^{\text{opt}} \in N \times \hat{\bandwidth}_{\text{data}}$ (Figure \ref{fig:overview}, bottom).
In order to prepare the data, a row of zeros was prepended and appended to each $A^{\text{opt}}$ to serve as the initial inputs and final outputs for the model.
Next, a column with an indicator for whether the row was the first or last was prepended.
This resulted in training data of the form:
\begin{equation}
    \label{eq:graphRNNdata}
    \begin{bmatrix}
        1 & 0 & \cdots & 0 \\
        0 & A^{\text{opt}}_{0, 0} & \cdots & A^{\text{opt}}_{0, \cmbandwidth_{\text{data}}} \\
        \vdots & \vdots & \ddots & \vdots \\
        0 & A^{\text{opt}}_{N, 0} & \cdots & A^{\text{opt}}_{N, \cmbandwidth_{\text{data}}} \\
        1 & 0 & \cdots & 0 \\
    \end{bmatrix}.
\end{equation}
In order to train the model, the data were placed into \texttt{PackedSequence} objects in PyTorch \citep{Paszke_PyTorch_An_Imperative_2019}, enabling batched training with variable sequence lengths.

\paragraph{GraphRNN architecture.}
The architecture is a two layer MLP followed by a four layer GRU followed by two layer MLP:
\begin{enumerate}
    \itemsep0em 
    \item[] \textbf{GraphRNN layers}
    \item \texttt{Linear}($\cmbandwidth_{\text{data}} + 1$ $\mapsto$ 128)
    \item \texttt{BatchNorm1D}
    \item \texttt{ReLU}
    \item \texttt{Linear}(128 $\mapsto$ 128)
    \item \texttt{GRU}(4 layers, 128 $\mapsto$ 128)
    \item \texttt{Linear}(128 $\mapsto$ 128)
    \item \texttt{BatchNorm1D}
    \item \texttt{ReLU}
    \item \texttt{Linear}(128 $\mapsto$ $\cmbandwidth_{\text{data}} + 1$)
\end{enumerate}

\paragraph{GraphRNN hyperoptimization.}
The GraphRNN hyperparameter ranges were:
\begin{itemize}
    \item Learning rate $\sim$ Log Uniform $[10^{-4}, 10^{-2}]$
    \item Weight decay $\sim$ Log Uniform $[10^{-5}, 10^{-1}].$
\end{itemize}

\paragraph{GraphRNN sampling.}
Rows of the data matrix constructed in Eq. \ref{eq:graphRNNdata} were sampled according to the logits $\ell$ output at the last layer of the model adjusted by a temperature parameter, $\tau$.
That yielded row $i$ edge probabilities:
\begin{equation}
    p[(v_i, v_j) \in \mathcal{E}] \sim \text{Bernoulli}\left(\frac{1}{\tau} \ell_{i, j + 1}\right).
\end{equation}
$\tau$ was selected for each model to minimize mean MMD on the validation data.
The sampling process was halted when an indicator was sampled.

\subsection{Graphite} \label{app:graphite-details}
Graphite \cite{pmlr-v97-grover19a} is a VAE adapted for graph data with one set of latent variables per node.
Graphite predicts edge probabilities using a pairwise kernel between node representations at the end of the decoder.
We kept the general design while making a few architectural changes for simplicity and performance.

\paragraph{Graphite data pre-processing.}
For every graph $G = \left(\mathcal{V}, \mathcal{E}\right)$ the node order was selected using either BFS (standard variation) or C-M (\shortmethodname variation).
In the C-M case, we also found the bandwidth resulting from the order $\cmbandwidth$.
We constructed node features $X \in \mathbb{R}^{N \times 16}$ using transformer-style positional encodings \cite{attention_vaswani}.
In the original work, \citet{pmlr-v97-grover19a} used one-hot positional encodings when there were no node features.
For each graph, an edge set was constructed for the decoder model.
In the BFS case, the edge set corresponding to a fully connected graph was used. 
In the C-M case, the edge set for a graph with bandwidth $\cmbandwidth$ was defined as in  Eq. \ref{eq:bandwidth_edgeset}.
\paragraph{Graphite architecture.}
The architecture was implemented using PyTorch Geometric \citep{Fey/Lenssen/2019}.
With respect to the original implementation, we used \texttt{GINE} layers \cite{Hu*2020Strategies}  rather than graph convolutions, and GELU activation \citep{hendrycks_gaussian_2020} rather than ReLU.
Each \texttt{GINE} layer contained a two-layer MLP with the following architecture:
\begin{enumerate}
    \itemsep0em 
    \item[] \texttt{GINE MLP} layers with hidden dimension $h$
    \item \texttt{Linear}($ h\mapsto 2h$)
    \item \texttt{BatchNorm1D}
    \item \texttt{GELU} 
    \item \texttt{Linear}($ 2h\mapsto h$)
    \item \texttt{BatchNorm1D}
    \item \texttt{GELU}.
\end{enumerate}
We made use of a stack of \texttt{GINE} layers, which we refer to as \ginestack (Algorithm \ref{alg:gine-stack}).
The Graphite encoder was a \ginestack with $h = 32$.
The variational marginals $\boldsymbol{\mu}, \boldsymbol{\sigma}$ (Eq. \ref{eq:graphite-mu-sigma}) were 32-dimensional and computed using a single linear layer each from the output of the encoder.
The decoder also employed a \ginestack with $h = 32$ with a few modifications~(Algorithm \ref{alg:graphite-decoder}).
The most consequential change in our experiments was replacing the final edge probability layer.
Rather than a dot product as in the original implementation, we observed better performance and lower variance with a two-layer MLP that takes as input concatenated pairs of node embeddings (Algorithm \ref{alg:graphite-decoder}, final three lines).

\begin{algorithm}
\caption{\ginestack}\label{alg:gine-stack}

\textbf{Inputs}: node features $X \in \mathbb{R}^{N \times d}$, edge list $\mathcal{E}$, edge features $E\in \mathbb{R}^{|\mathcal{E}| \times k}$ hidden dimension $h$\\
\textbf{Outputs}: updated node features $\hat{X}$\\
$X = \texttt{Linear}(d \mapsto h)(X)$\\
$X = \texttt{BatchNorm1D}(X)$\\
$X = \texttt{GELU}(X)$\\
$X_0 = \texttt{GINE}(X, \mathcal{E}, E)$\\
$X_1 = \texttt{GINE}(X_0, \mathcal{E}, E)$\\
$X_2 = \texttt{GINE}(X_1, \mathcal{E}, E)$\\
$\hat{X} = [X_0|X_1|X_2]$
\end{algorithm}

\begin{algorithm}
\caption{Graphite decoder. $\circ$ denotes function composition.}
\label{alg:graphite-decoder}
\textbf{Inputs}: node features $X \in \mathbb{R}^{N \times 16}$, embeddings $Z \in \mathbb{R}^{N \times 32}$, edge list $\mathcal{E}$, edge features $E\in \mathbb{R}^{|\mathcal{E}| \times k}$\\
\textbf{Outputs}: edge probability logits $\ell \in \mathbb{R}^{|\mathcal{E}|}$\\
$P = \texttt{Linear}(16 \mapsto 32)(X)$\\
$P = \texttt{GELU}(\texttt{BatchNorm1D})(P))$\\
$X_0 = Z + P$ \\
$X_1 = \text{\ginestack} (X_0, \mathcal{E}, E)$\\
$X_2 = [P|X_1]$\\
$X_3 = \texttt{GELU} \circ \texttt{BatchNorm1D} \circ \texttt{Linear}(112 \mapsto 32)(X_2)$\\
$K = [X|X_3]$ \\
$\ell_{i, j}^1 = \texttt{Linear}(32 \mapsto 1) \circ \texttt{GELU} \circ \texttt{Linear}(96 \mapsto 32) [K_i|K_j]$\\
$\ell_{i, j}^2 = \texttt{Linear}(32 \mapsto 1) \circ \texttt{GELU} \circ \texttt{Linear}(96 \mapsto 32) [K_j|K_i]$ \quad \# $j, i$ order to preserve symmetry\\
$\ell_{i, j} = \ell_{i, j}^1 + \ell_{i, j}^2$
\end{algorithm}

\paragraph{Graphite loss function.}
The loss was the standard VAE loss function with a hyperoptimized weight $\beta$ on the KL divergence term:
\begin{equation}
    \label{eq:graphite-loss}
    \mathcal{L}(\mathcal{E}, \ell, \boldsymbol{\mu}, \boldsymbol{\sigma}) = \frac{\beta}{|\mathcal{E}|} \sum_{i=1}^N\left(\boldsymbol{\mu}^2 + \boldsymbol{\sigma}^2 - \log[\boldsymbol{\sigma}] - \frac{1}{2} \right)_i + \frac{1}{|\mathcal{E}|}\texttt{BCE}(\ell, \mathcal{E}),
\end{equation}
where $\ell$ denotes the model predicted edge logits and \texttt{BCE} is the binary cross entropy.

\paragraph{Graphite hyperoptimization.}
The Graphite hyperparameter ranges were:
\begin{itemize}
    \item Learning rate $\sim$ Log Uniform $[10^{-4}, 10^{-2}]$
    \item KL-divergence weight ($\beta$) $\sim$ Log Uniform $[1, 10^{-5}].$
\end{itemize}

\paragraph{Graphite sampling.}

The latent variables $Z$ were sampled independently from the standard normal distribution.
Edge probabilities were then sampled from the decoder's edge probabilities (Eq. \ref{eq:graphite_P_E}, Algorithm \ref{alg:graphite-decoder}).
The number of nodes and bandwidths used to construct the decoder's message passing graph were sampled from the empirical distribution of the training data.

\subsection{EDP-GNN} \label{app:edpgnn-details}
EDP-GNN \cite{pmlr-v108-niu20a} is a permutation invariant score-based generative model for graphs.
\citet{pmlr-v108-niu20a} used annealed Langevin dynamics to sample from their model and used a variance schedule with six time steps.
We switched to the DDPM \citep{ddpm-ho-2020} framework, which we found led to more reliable results and faster sampling in preliminary experiments.

\paragraph{EDP-GNN data pre-processing.}
For every graph $G = \left(\mathcal{V}, \mathcal{E}\right)$ a node order was selected using either BFS (standard variation) or C-M (\shortmethodname variation).
In the C-M case, we also found the bandwidth $\cmbandwidth$.
We then constructed an edge set $\mathcal{E}'$ of edges not in the original graph, which the model was trained to distinguish from the real edges.
In the BFS case, these were all of the edges not in $\mathcal{E}$, i.e., $\mathcal{E}' = \{(i, j) \forall i \neq j\} - \edgeset$.
In the C-M case, these were the edges included in a graph with $\bandwidth(G) = \cmbandwidth$ (Eq. \ref{eq:bandwidth_edgeset}), and not in $\mathcal{E}$, that is, $\mathcal{E}' = \hat{\mathcal{E}} - \mathcal{E}$. 
Edge features $E \in \mathbb{R}^{|\edgeset| + |\edgeset'|}$ were constructed to encode whether each edge is in $\edgeset$  or $\edgeset'$ with 1 to indicate $\in \edgeset$ and -1 to indicate $\in \edgeset'$.
Node features $X \in \mathbb{R}^{N \times 16}$ were constructed using transformer-style positional encodings \cite{attention_vaswani}.
Time embedding features $T$ used for time conditioning were constructed using 128-dimensional positional encodings.

\begin{figure}[h]
\begin{center}
\centerline{
\includegraphics[width=0.85\textwidth]{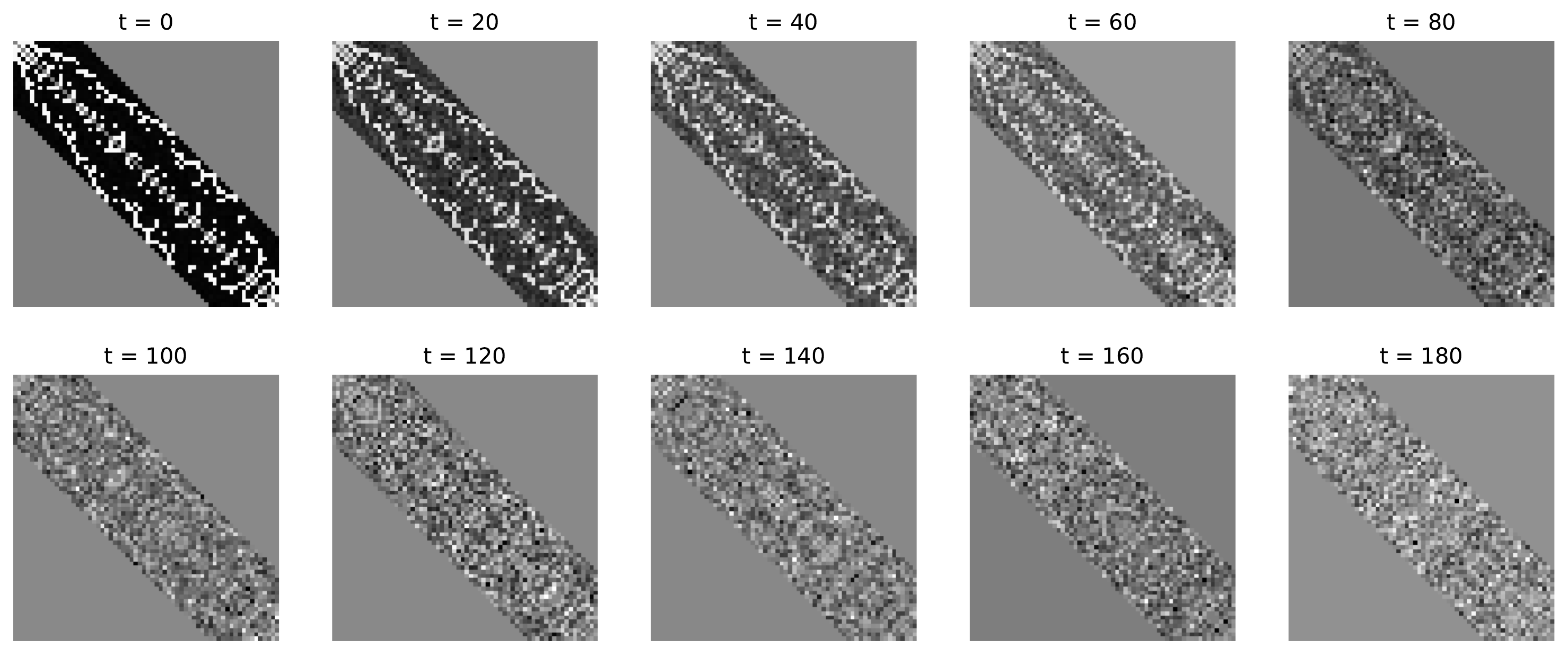}
}
\caption{Visualization of cosine variance schedule forward diffusion with 200 steps on a planar graph with restricted bandwidth.}
\label{fig:cosine-schedule}
\end{center}
\vskip -0.2in
\end{figure}

\paragraph{EDP-GNN diffusion hyperparameters.}
We used a cosine variance schedule \citep{nichol2021improved} with 200 steps.
The effect of this schedule on a restricted adjacency matrix of a planar graph is shown in Figure \ref{fig:cosine-schedule}.
We used the noise predicting parameterization $\boldsymbol{\epsilon}_\theta$ introduced by \citet{ddpm-ho-2020}.

\paragraph{EDP-GNN architecture.}
Our implementation of EDP-GNN was built using the previously introduced \ginestack (Algorithm \ref{alg:gine-stack}) followed by a two layer MLP operating on edge features pairs of node representations to predict the sampled noise $\boldsymbol{\epsilon}$.
The resultant architecture for the molecular datasets is shown in Algorithm $\ref{alg:edp-gnn}$. We used a node embedding size of 64 for the generic datasets instead of 128.

\begin{algorithm}
\caption{Modified EDP-GNN architecture. $\circ$ denotes function composition.}\label{alg:edp-gnn}
\textbf{Inputs}: node features $X \in \mathbb{R}^{N \times 16}$, edge list $\edgeset_\cup = \mathcal{E} \cup \edgeset'$, edge features $E\in \mathbb{R}^{|\edgeset_\cup|}$, time embeddings $T \in \mathbb{R}^{128}$\\
\textbf{Outputs}: noise predictions $\boldsymbol{\epsilon}_\theta \in \mathbb{R}^{|\mathcal{E} \cup \edgeset'|}$\\
$P = \texttt{GELU} \circ \texttt{Linear}(16 \mapsto 128)(X)$\\
$T_0 = \texttt{GELU} \circ \texttt{Linear}(128 \mapsto 128)(T)$
$X_0 = T_0 + P$ \\
$X_1 = \text{\ginestack} (X_0, \mathcal{E}_\cup, E)$\\
$X_2 = [P|T_0|X_1]$\\
$X_3 = \texttt{GELU} \circ \texttt{BatchNorm1D} \circ \texttt{Linear}(768 \mapsto 128)(X_2)$\\
$K = [X|X_3]$ \\
$E^0 = \texttt{GELU} \circ \texttt{Linear}(1 \mapsto 128)(E)$\\
$\epsilon_{i, j}^1 = \texttt{Linear}(128 \mapsto 1) \circ \texttt{GELU} \circ \texttt{Linear}(384 \mapsto 128) [K_i|K_j|E^0_{i, j}]$ \\
$\epsilon_{i, j}^2 = \texttt{Linear}(128 \mapsto 1) \circ \texttt{GELU} \circ \texttt{Linear}(384 \mapsto 128) [K_j|K_i|E^0_{j, i}]$ \quad \# $j, i$ order to preserve symmetry\\
$\boldsymbol{\epsilon}_{\theta, i, j} = \epsilon_{i, j}^1 + \epsilon_{i, j}^2$
\end{algorithm}

\paragraph{EDP-GNN hyperoptimization.}
The learning rate was hyperoptimized with a distribution $\sim$ Log Uniform $[10^{-4}, 10^{-2}]$.

\paragraph{EDP-GNN sampling.}
We used the DDPM sampling algorithm \cite{ddpm-ho-2020}.

\subsection{Likelihood Evaluation}

To compute the likelihood of test set graphs as an evaluation metric, we use the following strategies. 
For GraphRNN, we use the auto-regressive log-likelihood.
For Graphite, we use the variational ELBO as a lower bound.
For EDP-GNN, we use the ELBO for diffusion models, as shown in \cite{ddpm-ho-2020}.

\section{Datasets}\label{app:datasets}
All datasets were filtered so that there was one connected component per example.

\subsection{Example Datasets for Table \ref{tab:naturalBWs}}

All datasets, except Peptides-func, are available through the TUDataset collection \cite{morris2020tudataset}. Peptides-func is available in the Long Range Graph Benchmark \cite{dwivedi2022LRGB}.

\subsection{Generic Datasets}\label{app:generic-datasets}
\paragraph{Community2.} For each graph the number of nodes $N$ was sampled uniformly between 60 and 160.
Each community was then generated using an Erdos-Renyi model with edge probability 0.3.
Then edges between the two communities were sampled with probability 0.05.
Finally, the largest connected component of the resultant graph was selected.
\paragraph{Planar.}
For each graph 64 2D node coordinates were sampled uniformly between zero and one.
A Delaunay triangulation was performed on these coordinates.
Two nodes were considered adjacency if they shared a vertex in the triangulation.
\paragraph{Grid2d.} All unique pairs of side lengths between 10 and 20 were enumerated.
For each side length pair, a 2D grid graph was generated.
Since this yielded only 66 graphs, each graph in the training and validation sets were included five times with different random BFS and C-M orders each time.
\paragraph{DD.} The DD graphs were filtered so that each had between 100 and 500 nodes as in \cite{pmlr-v80-you18a}, going from 1178 to 918 graphs.
\paragraph{Enzymes.} The enzymes graphs were filtered so that each had $10 \leq N \leq 125$ going from 600 to 556 graphs.
\paragraph{Proteins.} The proteins graphs were filtered so that each had $10 \leq N \leq 125$ going from 1113 to 904 graphs.

\subsection{Molecular Datasets}\label{app:mol-datasets}
\paragraph{ZINC250k.} No filtering was required.
\paragraph{Peptides-func.} Removing graphs with more than one connected component filtered 15535 graphs down to 15375.

\section{Additional Experimental Results}
\subsection{MMD Metrics}
We include the individual MMDs results (summarized by the mean MMD in the main text) in \cref{tab:results-generic-extended}.

\subsection{Computational Metrics}

We include computational metrics for all datasets in \cref{tab:results-computational_extended}.

\subsection{Impact of Output Space Reduction to Performance Improvements}  \label{app:space-improvements}

\begin{figure}[h]
\vskip 0.1in
\begin{center}
\centerline{
\includegraphics[width=0.4\columnwidth]{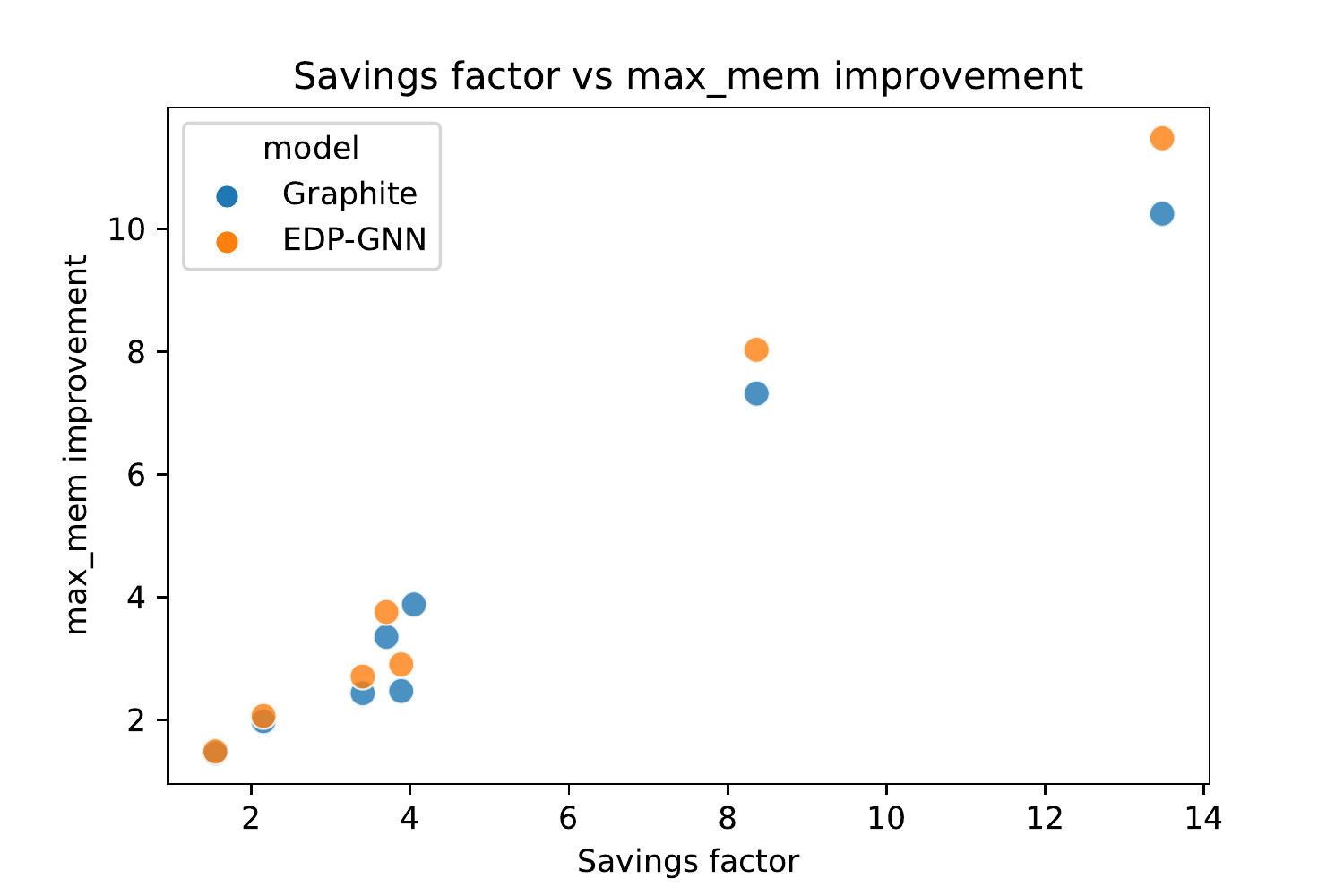}}
\caption{Savings factor versus improvement in memory usage with and without \shortmethodname for different datasets, for Graphite and EDP-GNN models.}
\label{fig:saving-vs-improvement}
\end{center}
\vskip -0.2in
\end{figure}

We study whether the theoretical improvement in space/time complexity given by \shortmethodname translates well into an empirical improvement in generation quality and computational complexity.
To do this, we analyze the relationship between the savings factor, which summarizes the space reduction given by the C-M bandwidth reparameterization (Table \ref{tab:naturalBWs}), and the performance improvement, measured as the ratio between standard models' metrics and their +\shortmethodname extensions.
Interestingly, we observe a high correlation (Spearman-r of 0.90 for Graphite, 0.89 for EDP-GNN, and 0.62 for GraphRNN) between the savings ratio and the log-likelihood improvement across the 8 datasets. Additionally, we observe a high correlation (Spearman-r of 0.97 for Graphite and 0.96 for EDP-GNN) between the savings ratio and the improvement in GPU memory usage (Figure~\ref{fig:saving-vs-improvement}) across the 8 datasets.
This analysis suggests that (1) the theoretical improvement correlates   with the empirical advantage, and (2) we can get an estimate of the expected empirical improvement for a specific dataset in advance, without actually training a model.

\section{Why Molecular Graphs Have Low Bandwidth?} \label{app:bw-molecular-theory}
In this section, we discuss the bandwidth of molecular graphs, i.e. graphs whose nodes and edges describe atoms and bonds, respectively.
As shown in Table~\ref{tab:naturalBWs}, all the considered molecular datasets have low average empirical bandwidth (in particular, the first section of Table~\ref{tab:naturalBWs} and Peptides-func correspond to molecular graphs). Additionally, the average bandwidth increases slowly as the average number of nodes $N$ in the datasets increases (for the considered datasets, the average bandwidth $\cmbandwidth$ varies between $2.8 \pm 0.7$ and $5.7 \pm 2.6$, while the average number of nodes $N$ varies between $10.1 \pm 0.7$ and $150.9 \pm 84.5$). Furthermore, given that the empirical bandwidth is estimated through a heuristic algorithm, part of the increase in $\cmbandwidth$ for higher $N$ can be explained by a slightly reduced algorithm efficacy for larger graphs.

All these empirical observations motivate more theoretical research on   \emph{why} all molecular graphs seem to have restricted bandwidth. To investigate this question, we have derived several upper bounds on the bandwidth of molecular graphs. These bounds show that, indeed, the inherent properties of molecular graphs confer low bandwidth. These results further strengthen the universal validity of \shortmethodname, beyond the datasets considered in this paper. 

Several bounds to the graph bandwidth for molecular graphs are discussed in the following:
\begin{description}
    \item[Molecules have planar graphs and bounded max degree.] As highlighted in the cheminformatics literature, molecules with non-planar graphs are extremely rare \cite{wester2008scaffold}. In practice, all molecules included in drug-like libraries (e.g., ZINC250k dataset) have planar graphs. Additionally, because of chemical bonding rules, all molecular graphs have a bounded maximum degree (4-6, depending on the atom types in the molecule). The combination of these two properties (bounded degree and planarity) guarantees sub-linear bandwidth, in $\bigO\left(\frac{N}{\log_{\Delta}\left(N\right)}\right)$, with $\Delta$ being the maximum degree, as proved by   \citet{bottcher2010bandwidth}.
    \item[Molecules as combination of motifs.] A less rigorous and more intuitive explanation comes from the fact that synthesizable molecules tend to consist of small components connected in a few (typically, one to three) places. Under the (simplified) assumption that a molecular graph is a string of connected components, the graph’s bandwidth is upper-bounded by the size of the largest component. This can be seen by considering the resulting block-diagonal adjacency matrix: each component corresponds to a block in the adjacency matrix, and the bandwidth of the graph corresponds to the bandwidth of the largest block/component. For example, in a linear molecule (like a long alkane chain), the bandwidth is one (independent of the length of the chain) because each carbon (individual node) is only directly connected to its neighbors.
    \item[Crystal structure as molecular upper bound.] Another  chemistry-inspired upper bound can be derived by considering regular chemical graphs with degree four. If we allow all 4-regular graphs, this is insufficient to constrain the bandwidth since random 4-regular graphs are expander graphs and have bandwidth in $\bigO\left(\frac{N}{\log\left(N\right)}\right)$. Instead, we consider the most densely packed form of carbon, diamond, which is formed by a 3D lattice. The graph of diamond’s structure is a subgraph of the 3D grid graph \cite{eppstein2009isometric}. Interestingly, the 3D grid graph has bandwidth that scales with the square root of the number of nodes \cite{otachi2011bandwidth}, providing a tighter upper bound in $\bigO\left(\sqrt{N}\right)$.
\end{description}

Further research on theoretical upper bounds on the bandwidth of graphs occurring in common domains, such as molecules and other biological objects, will be the subject of future work.

\begin{table*}[t]

  \caption{Graph generation results with individual MMD values. \textbf{Bold} indicates best results compared to the other model of the same type and dataset. Significance was determined by Welch's t-test with five replicates per model. Models are considered comparable when $p \ge 0.05$.  Graph statistics (degree, cluster, orbit, spectra) are reported as MMD$^{2}$. Mean computed across individual statistics for each model/dataset. OOM denotes out-of-memory issues. Hyphen (–) denotes not applicable metric/model.}
  
\label{tab:results-generic-extended}
\vskip 0.15in
\begin{center}
\begin{small}

\resizebox{\textwidth}{!}{
\renewcommand{\arraystretch}{1.1}
\begin{tabular}{llcccccccccccc}
\toprule
 &  & \multicolumn{4}{c}{\textsc{community2}} & \multicolumn{4}{c}{\textsc{planar}} & \multicolumn{4}{c}{\textsc{grid2d}} \\
\cmidrule(l{2pt}r{2pt}){3-6}
\cmidrule(l{2pt}r{2pt}){7-10}
\cmidrule(l{2pt}r{2pt}){11-14}
 &  & $\downarrow$ Deg. & $\downarrow$ Cluster & $\downarrow$ Orbit & $\downarrow$ Spectra & $\downarrow$ Deg. & $\downarrow$ Cluster & $\downarrow$ Orbit & $\downarrow$ Spectra & $\downarrow$ Deg. & $\downarrow$ Cluster & $\downarrow$ Orbit & $\downarrow$ Spectra \\
\midrule
\multirow{2}{*}{GraphRNN} & Standard & \textbf{0.016} & \textbf{0.046} & \textbf{0.024} & 0.024 & \textbf{0.056} & \textbf{0.309} & 0.617 & \textbf{0.080} & 0.403 & \textbf{0.000} & 0.714 & 0.174 \\
 & \shortmethodname [ours] & \textbf{0.034} & \textbf{0.041} & \textbf{0.017} & \textbf{0.006} & \textbf{0.060} & \textbf{0.311} & \textbf{0.481} & \textbf{0.079} & \textbf{0.037} & 0.797 & \textbf{0.066} & \textbf{0.061} \\
\midrule
\multirow{2}{*}{Graphite} & Standard & 0.146 & \textbf{0.047} & \textbf{0.015} & \textbf{0.011} & \textbf{0.289} & \textbf{0.304} & \textbf{0.749} & \textbf{0.104} & 0.500 & \textbf{1.300} & 0.601 & 0.198 \\
 & \shortmethodname [ours] & \textbf{0.114} & \textbf{0.047} & \textbf{0.015} & \textbf{0.013} & \textbf{0.311} & 0.323 & 1.110 & \textbf{0.128} & 0.069 & \textbf{1.970} & 0.038 & 0.035 \\
\midrule
\multirow{2}{*}{EDP-GNN} & Standard & \textbf{0.037} & \textbf{0.056} & \textbf{0.020} & \textbf{0.008} & \textbf{0.229} & \textbf{0.400} & \textbf{1.120} & \textbf{0.086} & \textbf{0.428} & \textbf{1.380} & \textbf{0.661} & \textbf{0.113} \\
 & \shortmethodname [ours] & \textbf{0.066} & \textbf{0.048} & \textbf{0.032} & \textbf{0.014} & \textbf{0.179} & \textbf{0.377} & \textbf{1.250} & \textbf{0.092} & \textbf{0.415} & \textbf{1.450} & \textbf{0.392} & \textbf{0.101} \\
\midrule
 &  & \multicolumn{4}{c}{\textsc{DD}} & \multicolumn{4}{c}{\textsc{ENZYMES}} & \multicolumn{4}{c}{\textsc{PROTEINS}} \\
\cmidrule(l{2pt}r{2pt}){3-6}
\cmidrule(l{2pt}r{2pt}){7-10}
\cmidrule(l{2pt}r{2pt}){11-14}
 &  & $\downarrow$ Deg. & $\downarrow$ Cluster & $\downarrow$ Orbit & $\downarrow$ Spectra & $\downarrow$ Deg. & $\downarrow$ Cluster & $\downarrow$ Orbit & $\downarrow$ Spectra & $\downarrow$ Deg. & $\downarrow$ Cluster & $\downarrow$ Orbit & $\downarrow$ Spectra \\
\midrule
\multirow{2}{*}{GraphRNN} & Standard & \textbf{0.066} & \textbf{0.155} & \textbf{0.410} & \textbf{0.065} & 0.011 & 0.045 & \textbf{0.021} & \textbf{0.018} & \textbf{0.004} & \textbf{0.040} & \textbf{0.015} & \textbf{0.010} \\
 & \shortmethodname [ours] & \textbf{0.092} & \textbf{0.229} & \textbf{0.489} & \textbf{0.125} & \textbf{0.003} & \textbf{0.039} & \textbf{0.010} & \textbf{0.014} & 0.012 & 0.045 & \textbf{0.011} & 0.013 \\
\midrule
\multirow{2}{*}{Graphite} & Standard & 0.316 & \textbf{0.316} & \textbf{0.656} & \textbf{0.186} & 0.204 & \textbf{0.046} & 0.099 & 0.078 & 0.293 & \textbf{0.096} & 0.111 & 0.114 \\
 & \shortmethodname [ours] & \textbf{0.239} & \textbf{0.245} & \textbf{0.492} & \textbf{0.118} & \textbf{0.042} & \textbf{0.039} & \textbf{0.052} & \textbf{0.018} & \textbf{0.037} & \textbf{0.043} & \textbf{0.052} & \textbf{0.016} \\
\midrule
\multirow{2}{*}{EDP-GNN} & Standard & OOM & OOM & OOM & OOM & \textbf{0.098} & \textbf{0.069} & 0.159 & \textbf{0.042} & 0.119 & 0.064 & 0.082 & 0.045 \\
 & \shortmethodname [ours] & \textbf{0.184} & \textbf{0.208} & \textbf{0.738} & \textbf{0.065} & \textbf{0.027} & \textbf{0.033} & \textbf{0.036} & \textbf{0.013} & \textbf{0.027} & \textbf{0.038} & \textbf{0.021} & \textbf{0.012} \\
\midrule
 &  & \multicolumn{4}{c}{\textsc{zinc250k}} & \multicolumn{4}{c}{\textsc{peptides-func}} \\
\cmidrule(l{2pt}r{2pt}){3-6}
\cmidrule(l{2pt}r{2pt}){7-10}
 &  & $\downarrow$ Deg. & $\downarrow$ Cluster & $\downarrow$ Orbit & $\downarrow$ Spectra & $\downarrow$ Deg. & $\downarrow$ Cluster & $\downarrow$ Orbit & $\downarrow$ Spectra \\
\midrule
\multirow{2}{*}{GraphRNN} & Standard & 0.025 & \textbf{0.045} & 0.012 & \textbf{0.071} & \textbf{0.009} & \textbf{0.004} & \textbf{0.000} & \textbf{0.108} \\
 & \shortmethodname [ours] & \textbf{0.011} & \textbf{0.044} & \textbf{0.005} & \textbf{0.057} & \textbf{0.008} & \textbf{0.001} & \textbf{0.001} & \textbf{0.123} \\
\midrule
\multirow{2}{*}{Graphite} & Standard & 0.049 & 0.516 & 0.005 & 0.044 & 0.169 & \textbf{0.235} & \textbf{0.030} & \textbf{0.293} \\
 & \shortmethodname [ours] & \textbf{0.009} & \textbf{0.307} & \textbf{0.002} & \textbf{0.019} & \textbf{0.056} & \textbf{0.216} & \textbf{0.011} & \textbf{0.198} \\
\midrule
\multirow{2}{*}{EDP-GNN} & Standard & 0.174 & \textbf{0.055} & 0.024 & 0.170 & 0.159 & \textbf{0.041} & 0.047 & 0.213 \\
 & \shortmethodname [ours] & \textbf{0.015} & 0.528 & \textbf{0.004} & \textbf{0.023} & \textbf{0.050} & 0.371 & \textbf{0.007} & \textbf{0.144} \\
\bottomrule
\end{tabular}
}

\end{small}
\end{center}
\end{table*}

\begin{table*}[t]  

\caption{Computational cost results. \textbf{Bold} indicates best results compared to the other model of the same type and dataset. Significance was determined by Welch's t-test with five replicates per model. Models are considered comparable when $p \ge 0.05$.}
  
\label{tab:results-computational_extended}
\vskip 0.1in
\begin{center}
\begin{small}

\resizebox{\textwidth}{!}{
\renewcommand{\arraystretch}{1.1}
\begin{tabular}{llcccccccccccc}
\toprule
 &  & \multicolumn{3}{c}{\textsc{community2}} & \multicolumn{3}{c}{\textsc{planar}} & \multicolumn{3}{c}{\textsc{grid2d}} & \multicolumn{3}{c}{\textsc{DD}} \\
\cmidrule(l{2pt}r{2pt}){3-5}
\cmidrule(l{2pt}r{2pt}){6-8}
\cmidrule(l{2pt}r{2pt}){9-11}
\cmidrule(l{2pt}r{2pt}){12-14}
 &  & $\downarrow$ sample (s) & $\downarrow$ mem. (GB) & $\downarrow$ batch (s) & $\downarrow$ sample (s) & $\downarrow$ mem. (GB) & $\downarrow$ batch (s) & $\downarrow$ sample (s) & $\downarrow$ mem. (GB) & $\downarrow$ batch (s) & $\downarrow$ sample (s) & $\downarrow$ mem. (GB) & $\downarrow$ batch (s) \\
\midrule
\multirow{2}{*}{GraphRNN} & Standard & \textbf{0.673} & \textbf{0.059} & \textbf{0.008} & \textbf{0.095} & \textbf{0.036} & \textbf{0.007} & \textbf{0.490} & \textbf{0.115} & \textbf{0.016} & \textbf{1.470} & \textbf{0.142} & \textbf{0.015} \\
 & \shortmethodname [ours] & \textbf{0.538} & \textbf{0.061} & \textbf{0.009} & \textbf{0.094} & \textbf{0.036} & \textbf{0.007} & \textbf{0.520} & \textbf{0.112} & \textbf{0.018} & \textbf{1.500} & \textbf{0.149} & \textbf{0.016} \\
\midrule
\multirow{2}{*}{Graphite} & Standard & 1.460 & 0.805 & 0.012 & 0.513 & 0.266 & \textbf{0.010} & 6.400 & 3.270 & 0.022 & 10.90 & 5.870 & 0.028 \\
 & \shortmethodname [ours] & \textbf{1.050} & \textbf{0.553} & \textbf{0.010} & \textbf{0.242} & \textbf{0.135} & \textbf{0.009} & \textbf{0.842} & \textbf{0.447} & \textbf{0.012} & \textbf{2.380} & \textbf{1.510} & \textbf{0.013} \\
\midrule
\multirow{2}{*}{EDP-GNN} & Standard & 12.50 & 1.600 & 0.006 & 4.300 & 0.526 & \textbf{0.004} & 50.80 & 6.730 & 0.016 & OOM & OOM & OOM \\
 & \shortmethodname [ours] & \textbf{8.410} & \textbf{1.080} & \textbf{0.005} & \textbf{2.260} & \textbf{0.255} & \textbf{0.003} & \textbf{6.550} & \textbf{0.837} & \textbf{0.004} & \textbf{19.90} & \textbf{2.620} & \textbf{0.007} \\
\midrule
 &  & \multicolumn{3}{c}{\textsc{ENZYMES}} & \multicolumn{3}{c}{\textsc{PROTEINS}} & \multicolumn{3}{c}{\textsc{zinc250k}} & \multicolumn{3}{c}{\textsc{peptides-func}} \\
\cmidrule(l{2pt}r{2pt}){3-5}
\cmidrule(l{2pt}r{2pt}){6-8}
\cmidrule(l{2pt}r{2pt}){9-11}
\cmidrule(l{2pt}r{2pt}){12-14}
 &  & $\downarrow$ sample (s) & $\downarrow$ mem. (GB) & $\downarrow$ batch (s) & $\downarrow$ sample (s) & $\downarrow$ mem. (GB) & $\downarrow$ batch (s) & $\downarrow$ sample (s) & $\downarrow$ mem. (GB) & $\downarrow$ batch (s) & $\downarrow$ sample (s) & $\downarrow$ mem. (GB) & $\downarrow$ batch (s) \\
\midrule
\multirow{2}{*}{GraphRNN} & Standard & \textbf{0.129} & \textbf{0.019} & \textbf{0.011} & \textbf{0.333} & \textbf{0.023} & \textbf{0.007} & \textbf{0.080} & \textbf{0.014} & \textbf{0.004} & \textbf{7.690} & \textbf{0.082} & \textbf{0.014} \\
 & \shortmethodname [ours] & \textbf{0.239} & \textbf{0.019} & \textbf{0.014} & \textbf{0.533} & \textbf{0.021} & \textbf{0.009} & \textbf{0.080} & \textbf{0.014} & \textbf{0.004} & \textbf{7.680} & \textbf{0.080} & \textbf{0.013} \\
\midrule
\multirow{2}{*}{Graphite} & Standard & \textbf{0.450} & 0.085 & \textbf{0.014} & 0.227 & 0.105 & \textbf{0.008} & \textbf{0.071} & 0.040 & \textbf{0.008} & 3.520 & 1.740 & 0.015 \\
 & \shortmethodname [ours] & \textbf{0.050} & \textbf{0.035} & \textbf{0.010} & \textbf{0.056} & \textbf{0.031} & \textbf{0.010} & \textbf{0.083} & \textbf{0.016} & \textbf{0.010} & \textbf{0.268} & \textbf{0.170} & \textbf{0.008} \\
\midrule
\multirow{2}{*}{EDP-GNN} & Standard & 1.590 & 0.178 & \textbf{0.004} & 2.040 & 0.211 & \textbf{0.003} & 1.620 & 0.145 & \textbf{0.003} & 69.80 & 6.280 & 0.024 \\
 & \shortmethodname [ours] & \textbf{0.981} & \textbf{0.066} & \textbf{0.003} & \textbf{1.040} & \textbf{0.056} & \textbf{0.003} & \textbf{0.809} & \textbf{0.050} & \textbf{0.004} & \textbf{4.880} & \textbf{0.547} & \textbf{0.004} \\
\bottomrule
\end{tabular}
}

\end{small}
\end{center}
\end{table*}
\end{document}